\documentclass[10pt,journal,compsoc]{IEEEtran}

\usepackage{algorithm}
\usepackage{algorithmic}
\usepackage{subfigure}
\usepackage{graphicx}
\usepackage{amsmath}
\usepackage{amsthm}
\usepackage{booktabs}
\usepackage{multirow}
\usepackage{color}
\usepackage{url} 
\usepackage{hyperref}
\usepackage{footnote}
\usepackage{amsfonts}       
\usepackage{nicefrac}       
\usepackage{microtype}      
\usepackage{xcolor}         
\usepackage{diagbox}
\usepackage{wrapfig,lipsum}
\usepackage{amssymb,mathtools}
\usepackage{blindtext}
\usepackage{makecell}
\usepackage{longtable}
\usepackage{tabu}
\usepackage{enumitem}
\usepackage{subfigure}

\newcommand{\etal}{\emph{et al.}}

\ifCLASSOPTIONcompsoc
  \usepackage[nocompress]{cite}
\else
  \usepackage{cite}
\fi

\ifCLASSINFOpdf
\else
\fi

\begin{document}
\title{A Survey on Non-Autoregressive Generation \\ for Neural Machine Translation and Beyond}

\author{Yisheng~Xiao*,
        Lijun~Wu*,
        Junliang~Guo,
        Juntao~Li,
        Min~Zhang,
        Tao~Qin,~\IEEEmembership{Senior Member,~IEEE}
        and~Tie-Yan~Liu,~\IEEEmembership{Fellow,~IEEE}
\IEEEcompsocitemizethanks{
\IEEEcompsocthanksitem Yisheng Xiao, Juntao Li, and Min Zhang are with Soochow University, Suzhou, China. 
E-mail: ysxiaoo@stu.suda.edu.cn; ljt@suda.edu.cn; minzhang@suda.edu.cn.
\IEEEcompsocthanksitem Lijun Wu, Junliang Guo, Tao Qin, and Tie-Yan Liu are with Microsoft Research, Beijing 100080, China.
E-mail: lijuwu@microsoft.com; junliangguo@microsoft.com; taoqin@microsoft.com;
tyliu@microsoft.com.
}
\thanks{Yisheng Xiao and Lijun Wu contributed equally to this paper. Juntao Li is the corresponding author. This work is supported by NSFC No. 62206194 and the Beijing Academy of Artificial Intelligence.}
}

\markboth{Journal of \LaTeX\ Class Files,~Vol.~00, No.~0, April~2023}%
{Yisheng Xiao \MakeLowercase{\textit{et al.}}: A Survey on Non-Autoregressive Generation}
%

\IEEEtitleabstractindextext{%
\begin{abstract}
Non-autoregressive (NAR) generation, which is first proposed in neural machine translation (NMT) to speed up inference, has attracted much attention in both machine learning and natural language processing communities. 
While NAR generation can significantly accelerate inference speed for machine translation, the speedup comes at the cost of sacrificed translation accuracy compared to its counterpart, autoregressive (AR) generation. 
In recent years, many new models and algorithms have been designed/proposed to bridge the accuracy gap between NAR generation and AR generation. 
In this paper, we conduct a systematic survey with comparisons and discussions of various non-autoregressive translation (NAT) models from different aspects. Specifically, we categorize the efforts of NAT into several groups, including data manipulation, modeling methods, training criterion, decoding algorithms, and the benefit from pre-trained models. 
Furthermore, we briefly review other applications of NAR models beyond machine translation, such as grammatical error correction, text summarization, text style transfer, dialogue, semantic parsing, automatic speech recognition, and so on.
In addition, we also discuss potential directions for future exploration, including releasing the dependency of KD, reasonable training objectives, pre-training for NAR, and wider applications, etc.
We hope this survey can help researchers capture the latest progress in NAR generation, inspire the design of advanced NAR models and algorithms, and enable industry practitioners to choose appropriate solutions for their applications.
The web page of this survey is at \url{https://github.com/LitterBrother-Xiao/Overview-of-Non-autoregressive-Applications}.

\end{abstract}
}

\maketitle

\IEEEdisplaynontitleabstractindextext

\IEEEpeerreviewmaketitle

\IEEEraisesectionheading{\section{Introduction}\label{sec:introduction}}
\IEEEPARstart{M}achine translation~\cite{somers1992introduction} is one of the most critical and challenging tasks in natural language processing (NLP), which aims to translate natural language sentences from the source language to the target language.  
Recently, with the breakthrough of deep learning~\cite{lecun2015deep}, Neural Machine Translation (NMT)~\cite{bahdanau2015neural,cho2014learning,sutskever2014sequence,wu2016google,luong2015effective,sennrich2016neural}, which takes the different neural networks as backbone models, e.g., RNN~\cite{mikolov2010recurrent,bahdanau2015neural} and CNN~\cite{lecun1995convolutional,gehring2017convolutional}, has achieved outstanding performances, especially for the self-attention~\cite{lin2017structured} based Transformer~\cite{vaswani2017attention} models~\cite{dehghani2018universal,wu2019depth}.
NMT usually adopts the autoregressive generation (AR) method for translation (AT), which means the target tokens are one-by-one generated in a sequential manner. Therefore, AT is quite time-consuming when generating target sentences, especially for long sentences.
To alleviate this problem and accelerate decoding, non-autoregressive generation (NAR) for machine translation (NAT) is first proposed in~\cite{gu2018non},
which can translate/generate all the target tokens in parallel. Therefore, the inference speed is hugely increased, and much attention to NAT/NAR methods has been attracted with impressive progress~\cite{wang2018semi,ghazvininejad2019mask,xu2021does,ding2020understanding,guo2021self,wang2019non,guo2019non}.
However, the translation accuracy is damaged and sacrificed as a result of parallel decoding. Compared with AT, the tokens are generated without internal dependency for NAT models, unlike the AT models where the $t$-th token has previous $t-1$ contextual token information to help its generation.
Hence, the NAT models seriously suffer from lacking target side information to make predictions (e.g., decoding length) and correctly generate target translations. 
In summary, we attribute the main challenge of NAT models to the \textit{`failure of capturing the target side dependency.'}

\begin{figure*}[htp] 
\centering
\includegraphics[scale=0.585]{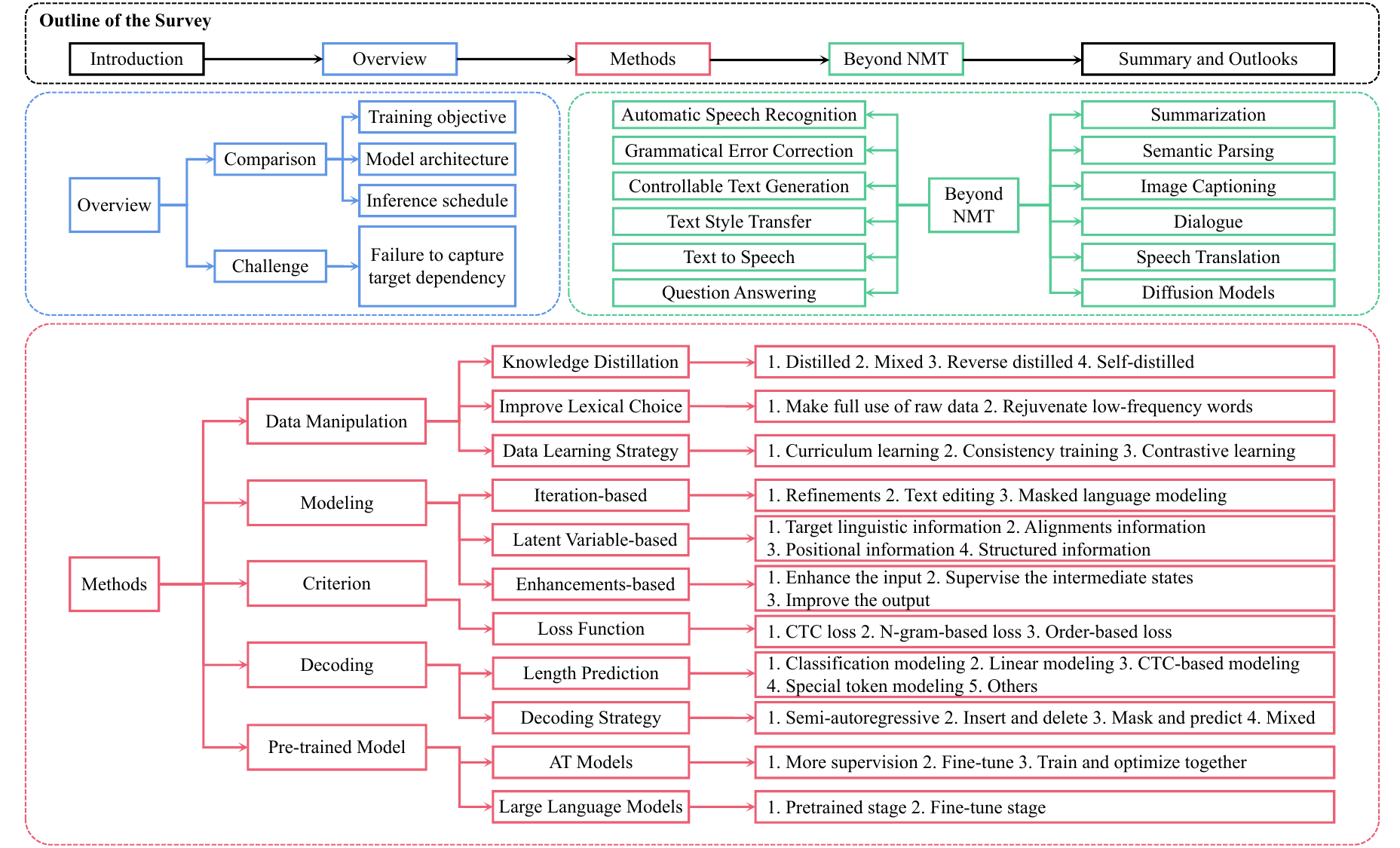}
\caption{Outline of the survey. We first review the developments of neural machine translation and non-autoregressive related methods. Then we present an overview of recent AT and NAT models, including a comparison of three different aspects (i.e., training objective, model architecture, and inference schedule). Besides, we also summarize the main challenge that recent NAT models encounter compared with AT models. To solve this challenge, we introduce several widely used methods to help improve the ability of NAT models at different levels, including data manipulation, modeling, criterion, decoding, and benefiting from pre-trained models. Then we present a summary of the above methods for the machine translation task. We also extend this survey to other extensive applications, such as automatic speech recognition, controllable text generation, question answering, image captioning, text summarization, grammatical error correction, text style transfer, dialogue, semantic parsing, text to speech, speech translation and diffusion models. Finally, some open problems and future outlooks are discussed. Best viewed this outline in color.}
\label{fig:Overview} 
\end{figure*}

To mitigate the above-mentioned challenge, significant efforts have been paid in the past few years from different aspects, e.g., data manipulation~\cite{ding2020understanding,guo2020fine}, modeling methods~\cite{sun2019fast,ghazvininejad2019mask}, decoding strategies~\cite{geng2021learning,graves2006connectionist}, to better capture the dependency on target side information. 
Although impressive progress has been achieved and the translation accuracy is greatly improved for NAT models, the translation quality still falls behind their AT counterparts. 
To continue narrowing the performance gap and facilitating the development of NAT in the future, a solid review of current NAT research is necessary. Therefore, we make the first comprehensive survey of existing non-autoregressive technologies for NMT in this paper. 
Our review summarizes the core challenge of NAT research and presents various advanced approaches to solve the challenge.
Specifically, we introduce the approaches from the following aspects:
\begin{itemize}[leftmargin=*]
    \setlength{\itemsep}{0pt}
    \setlength{\parskip}{0pt}
    \item \textbf{Data Manipulation.} As a data-driven task, the scale and quality of training data are crucial for NMT tasks. Due to the lack of target dependency for NAT models, lots of methods are proposed to reduce the complexity of the training data to provide an easier training task for them. 
    \item \textbf{Modeling.} Various advanced model architectures are proposed to better capture the target dependency, including iteration-based methods that can provide partially observed target information, latent variable-based methods that introduce latent variables to learn the target side dependency, and enhancements-based methods that directly provide stronger target side information to the input/output/intermediate states of the decoder.
    \item \textbf{Criterion.} Some works point out that the traditional cross-entropy loss is not optimal for NAT models and propose better criteria to improve the training of NAT models, including Connectionist Temporal Classification (CTC) based, N-gram-based, and order-based loss functions.
    \item \textbf{Decoding.} Decoding algorithm is another decisive factor in NMT models. Upon NAT models, different tricks are proposed to improve the decoding process and provide better translation results.
    \item \textbf{Pre-Trained Model.} Finally, given the strong representation capacity of pre-trained models, it is appealing to utilize them to improve the performance of NAT models. Therefore, lots of methods have been proposed to leverage the information from pre-trained AT models or large-scale language models to help the training of NAT models.
\end{itemize}
Besides summarizing the improving works for NAT, we also review other applications of NAR methods beyond NMT, such as automatic speech recognition, controllable text generation, question answering, image captioning, text summarization, grammatical error correction, text style transfer, dialogue, semantic parsing, text to speech, speech translation and diffusion models.
We further point out the recent trends and possible promising directions for future development, such as releasing the dependency of knowledge distillation, designing more reasonable training objectives,
and pre-training for NAR models.
We hope this survey paper can provide researchers and engineers with valuable insights to attract more people to either promote the development of NAT/NAR techniques or bring NAT/NAR methods into other fields, such as NAR text generation with large-scale pre-trained language models.
Besides, the up-to-the-minute solutions for each NAT problem and thorough analysis of model performance and computational cost are also expected to assist considerable industry practitioners.

The organization of this survey paper is as follows. To begin with, we make a comparison between AT and NAT models from different views, such as their training objectives, model architectures, and inference schedules. Then we analyze the main challenge that the current NAT models encounter in Section ~\ref{sec:overview}. We attribute the low quality of NAT models to the failure to well capture the target dependency, and from Section~\ref{sec:data} to Section~\ref{sec:benefition}, we summarize the efforts paid for improvement from different aspects, including data manipulations (Section~\ref{sec:data}), modeling methods (Section~\ref{sec:model}), training criterion (Section~\ref{sec:criterion}), decoding ways (Section~\ref{sec:decoding}), and the benefit from pre-trained models (Section~\ref{sec:benefition}). In addition, we also give a short summary of current NAT models in Section~\ref{sec:summary}.
Next, we investigate the extension of NAR generation approaches to other various applications beyond NMT in Section~\ref{sec:extension}.
At last, we conclude this paper and discuss our future outlooks in Section~\ref{sec:conclusion}. A detailed version of our survey outline is also shown in Figure~\ref{fig:Overview}.

\section{Overview of AT and NAT Models}
\label{sec:overview}
In this section, we first give an overall introduction to AT and NAT models. We briefly compare them from several different aspects, including the training objective, model architecture, and inference strategy. Besides, we also analyze the main challenge that NAT models encounter compared with AT.

\subsection{Comparison}
\label{sec:comparison}
Both AT and NAT models aim to make a correct sentence translation from a source language to a target language. Due to their unique characteristics, their differences are apparent in training, modeling, and inference. Before a detailed comparison, we first introduce the necessary notations. 

Given a dataset $D=\{(X, Y)_i\}_{i=1}^N$, where $(X, Y)_i$ refers to a paired sentence data, and $N$ is the size of the dataset. $X$ is the sentence to be translated from source language $\mathcal{X}$ and $Y$ is the ground-truth sentence from target language $\mathcal{Y}$. The goal of NMT models is to learn a mapping function $f(\cdot)$ from the source sentence to the target sentence $f: X\to Y$ to estimate the unknown conditional distribution $P(Y|X; \theta)$, where $\theta$ denotes the parameter set of a network model. We now compare the details of AT and NAT models as below.

\noindent\textbf{Training Objective.}
(1) For paired sentences $(X, Y)$, where $X=\{x_1, x_2, ..., x_{T_X}\}$ and $Y=\{y_1, y_2, ..., y_{T_Y}\}$, the training objective $\mathcal{L}_{\text{AT}}$ of an autoregressive NMT (AT) model is to maximize the following likelihood:
\begin{equation}
\mathcal{L}_{\text{AT}}=\sum\nolimits_{t=1}^{T_Y} \log P(y_t|y_{<t},X;\theta),
\end{equation}
where $y_t$ is the token to be translated at current time step $t$ and $y_{<t}$ are the tokens predicted in previous $t-1$ decoding steps. 
From the above equation, we can clearly see that the training of AT models adopts the autoregressive factorization in a left-to-right manner. Note that during training, the ground-truth target tokens are leveraged with the teacher forcing method~\cite{kolen2001field,vaswani2017attention}. In this way, the translation quality is guaranteed with the help of contextual dependencies.  

(2) In contrast, the non-autoregressive NMT (NAT) models~\cite{gu2018non} use the conditional independent factorization for prediction, and the objective is to maximize the likelihood:
\begin{equation}
\mathcal{L}_{\text{NAT}}=\sum\nolimits_{t=1}^{T} \log P(y_t|X;\theta),  
\end{equation}
notice that $T$ is the length of the target sentence. During training, $T=T_Y$ is the length of the ground-truth target sentence, while in inference, $T=P_L(X)$ which is usually predicted by a length prediction module $P_L$. Compared with AT models, it is obvious that the conditional tokens $y_{<t}$ are removed for NAT models. Hence, we can do parallel translations without autoregressive dependencies, and the inference speed is greatly improved. 

(3) Besides the AT and NAT models, researchers aim to find an intermediate state between current AT and NAT, which can also serve as a universal formulation of both models to achieve a balance between decoding speed and translation quality. For example, Wang~\etal~\cite{wang2018semi} propose a semi-autoregressive NMT (SAT) model, which keeps the autoregressive property in global but relieves it in local. Shortly speaking, SAT models can produce multiple target tokens in parallel at each decoding step (local non-autoregressive) and dependently generate tokens for the next step (global autoregressive). Mathematically, SAT models aim to maximize the following likelihood:
\begin{equation}
    \mathcal{L}_{\text{SAT}}=\sum\nolimits_{t=1}^{[(T-1)/k]+1} \log P(G_t|G_{<t},X;\theta),
\end{equation}
where $k$ denotes the number of the tokes that the SAT models parallelly generate at one time step. $G_t$ is a group of $k$ target tokens at $t$-th step. $G_{<t}$ is the $t-1$ groups of target tokens generated in the previous $t-1$ decoding steps.
Note that if $k=1$, it equals an AT model, and if $k=T$, it generalizes to a NAT model.

(4) In comparison, iteration-based NAT models share a spirit of mixed autoregressive and non-autoregressive translation, but on the sentence level with a refinement approach. 
That is, iteration-based NAT models keep the non-autoregressive property in every iteration step and refine the translation results during different iteration steps~\cite{lee2018deterministic,ghazvininejad2019mask}. 
The training goal is to maximize:
\begin{equation}
    \mathcal{L}_{\text{Iter}}=\sum\nolimits_{y_t \in Y_{tgt}} \log P(y_t|\hat{Y},X;\theta),
\end{equation}
where $\hat{Y}$ indicates the translation result of the last iteration, 
and $Y_{tgt}$ is the target of this iteration.

In the first iteration, only $X$ is fed into the model, which is the same as NAT models. After that, each iteration takes the translation generated from the last iteration as context for refinement to decode the translation. 
Generally speaking, NAT models with iterative refinements are viewed as iteration-based NAT models, while models with only one decoding step are viewed as fully NAT models.

\noindent\textbf{Model Architecture.}
As for model architecture, both AT and NAT models take the encoder and decoder framework for translation. The encoder and decoder can be different neural networks, such as RNN~\cite{mikolov2010recurrent}, CNN~\cite{gehring2017convolutional}, and Transformer~\cite{vaswani2017attention}. Due to the superior performance of the Transformer network, we focus on the Transformer model for discussion in this survey. The encoder is used to encode the source sentences, while the decoder is utilized for decoding the target sentence. 
Compared to AT and NAT models, they adopt the same encoder architecture, and the differences are reflected in the decoders to match the specific training objective. (1) Specifically, AT models need to prevent earlier decoding steps from peeking at information from later steps. Therefore, the constraint of an autoregressive factorization of the output distribution is required, and they adopt the strict causal mask by applying a lower triangular matrix in the self-attention module of the conventional Transformer decoder~\cite{vaswani2017attention}. (2) However, for NAT models, including the iteration-based NAT models, this constraint is no longer necessary, so they adopt the unmasked self-attention over all target tokens~\cite{gu2018non}. (3) As for SAT models, they adopt a coarse-grained lower triangular matrix as the causal mask, which means that they allow $k$ tokens to peep later information in the same group while keeping the constraint between different groups.

\noindent\textbf{Inference Schedule.}
When going to the inference stage, the differences are as follows. (1) The AT models predict the target tokens in a one-by-one manner, and the tokens predicted previously are fed back into the decoder to generate the next token. (2) While SAT models predict a group of target tokens at one time, the previously generated groups of tokens are fed into the decoder to generate the next group of tokens, which is the same as the AT models.
(3) For iteration-based NAT models, it needs $k$ iterations for inference. The translated results of the previous iteration will be fed into the decoder again for refinements. 
(4) As for fully NAT models, they generate all predicted target tokens at only one step, which greatly speeds up inference.
It is worth noting that AT and SAT models suffer from the gap between training and inference~\cite{zhang2019bridging,ranzato2016sequence,wu2018study}. That is, they utilize ground-truth target tokens during training, while the models can only take previously generated target tokens for inference. This indeed leads to inconsistency between training and inference and hence hurts the performance.
In contrast, fully NAT models are free from this trouble, but for iteration-based NAT models, prediction in the previous iteration is adopted for refinements, and this mismatched problem may be more serious. More details about this will be discussed in Section~\ref{sec:decoding}.

\subsection{The Main Challenge of NAT Models}\label{sec:main_challenge}
When achieving parallel decoding, a critical issue of NAT models is that they have no tokens with target information fed into the decoder~\cite{gu2018non} during training and inference. They can only rely on the source side information, which heavily increases the difficulty for NAT models. 
Previously, when Gu~\etal~\cite{gu2018non} first propose their NAT model, they notice that using nothing or only position embeddings in the first decoder layer results in poor translation performance. To alleviate this problem, they propose an initial module by copying the source tokens as the initialization for the decoder input. However, the source and target sentences from distinct languages are indeed different. This way does not help the decoder since no target information is given.

As a result, missing the target information leads the NAT models \textit{to fail to capture the target dependency of target tokens}, and we attribute the main challenge of low quality for NAT models to this defect. To better understand and further release this problem, we now give specific analysis with examples and also briefly show improvement methods in the following contents.

\label{subsec:problem}
\noindent\textbf{Understanding the Problem.}
Since no target information is fed into the decoder, NAT models remove the word dependency of the target sentence completely and generate target tokens entirely depending on the source sentence. 
Hence, terrible situations can happen to harm the translation quality. 
(1) First, the conditional independence assumption prevents a model from properly capturing the highly multi-modal distribution of target translations, which is called multi-modality problem~\cite{gu2018non}. Almost all the NAT models suffer from this trouble. Due to the strong assumption that each target token is predicted independently, if there are several different target sentences that can be viewed as reasonable translations, NAT models are possible to select fragments of each sentence and combine them as a candidate translation. Take an example, when translating \texttt{thank you} into German, \texttt{Vielen Dank} and \texttt{Danke} are both reasonable translations. However, NAT models may generate \texttt{Danke Dank}, which is truly unreasonable but should be impossible in AT models. 
Zhang~\etal~\cite{zhang2022study} also focus on the multi-modality problem but especially on the syntactic granularity. They first categorize the syntactic multi-modality problem into long-range and short-range types. Then they conduct a systematic study to evaluate the effectiveness of different loss functions for each kind. Finally, they introduce Combined CTC and O\scriptsize{A}\normalsize XE (CoCO) loss to alleviate the complicated syntactic multi-modality problem.
(2) Over-translation and under-translation~\cite{wang2019non} are also common translation errors. The issue of over-translation refers to the same word token being successively generated multiple times, leading the same token from different reasonable translations to appear at different positions in the final translation.  
The under-translation indicates that several necessary tokens in the source sentence are neglected, leading to several tokens missing in the translation results. 
Take an example, when translating German sentence \texttt{es gibt heute viele Farmer mit diesem Ansatz} into English sentence, a reasonable translation can be \texttt{there are lots of farmers doing this today}. However, NAT models may miss the word \texttt{of} (under-translation) or generate the word \texttt{of} twice (over-translation), leading the results to be \texttt{there are lots farmers doing this today} or \texttt{there are lots of of farmers doing this today}. This seriously harms the translation quality.
Instead, if target dependency is given as AT models, the problem of repetitive tokens and missing tokens can be avoided.

\begin{figure*}[t] 
\centering
\includegraphics[scale=0.88]{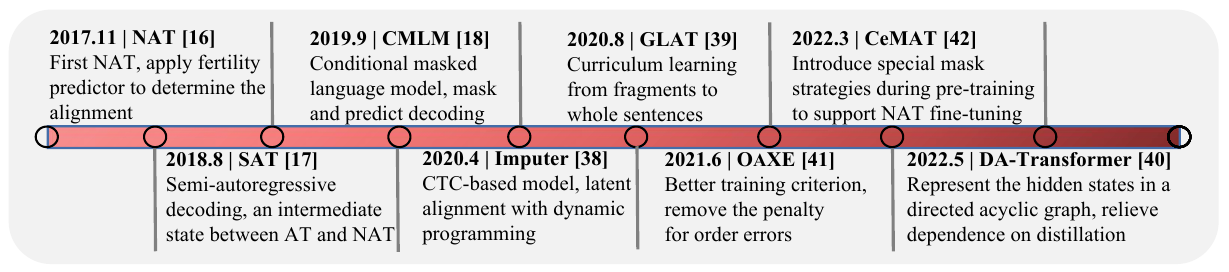}
\vspace{0.03cm}
\caption{Representative methods along the development of NAT models.}
\label{fig:TIMEdev} 
\end{figure*}

\subsection{Overview of Improving Methods}
As we discussed that NAT models are hard to model the target side dependency, various methods have been proposed to alleviate this problem by reducing the dependency of target tokens at different levels, which hence improves the ability of NAT models. 
Specifically, these methods include (1) data manipulation, which focuses on the improvements of training data corpus and data learning strategies, (2) improvements on the modeling level, where we first summarize two popular and widely used training frameworks (iteration-based methods and latent variable-based methods) along with various specific implementations of them. Besides, various other enhancements-based methods are introduced for NAT models, (3) improvements on the training criterion, where better criteria compared with traditional cross-entropy loss are proposed to meet the unique characteristics of NAT models, 
(4) improvements on the decoding level, where we introduce the tremendous progress made on length prediction and decoding strategy, and (5) benefiting from pre-trained models, i.e., guiding NAT models to benefit from their AT counterparts and other large-scale pre-trained language models such as BERT~\cite{kenton2019bert}. 
We plot a figure in the Appendix (Figure~\ref{fig:MODEL}) to structure these methods better.

In Table~\ref{tab:overview}, we give a summary and overview of different NAT models based on the above improvement category. In each category, we also present the specific sub-topics to better classify the category, along with the representative published works. 
For example, the decoding strategies can be divided to semi-autoregressive decoding (i.e., SAT~\cite{wang2018semi}), insert and delete (i.e., LevT~\cite{gu2019levenshtein}), mask and predict (i.e., Easy-First~\cite{kasai2020non}), and also mixed decoding (Unified~\cite{tian2020train}). 
Besides, for each work, we summarize a short description and list its published place (e.g., ACL, EMNLP), the decoding speed, and the performance on the mostly evaluated dataset WMT14 English$\to$German (EN$\to$DE) for a quick understanding.

Before introducing these methods, in Figure~\ref{fig:TIMEdev}, the most important and popular works along the NAT development are shown in the timeline. The NAT is first proposed in November 2017, and the inference speed is hugely improved, but the accuracy is far behind the AT model. After its birth, SAT~\cite{wang2018semi} is proposed to serve as the bridge between AT and NAT models with better translation performance. 
Other representative works are then introduced, including iteration-based methods: CMLM~\cite{ghazvininejad2019mask} and Imputer~\cite{saharia2020non}, fully NAT models: GLAT~\cite{qian2020glancing} and DA-Transformer~\cite{huang2022directed}.
These models mainly conduct improvements to the model structure. 
Besides, improvements based on training criteria are also introduced in O\scriptsize{A}\normalsize XE~\cite{du2021order}. 
Recently, CeMAT~\cite{li2022universal} is proposed to explore the potential of pre-training a non-autoregressive model and then fine-tuning on the translation task.
With the rapid growth of NAT models, their performance gap with AT models is narrowing, and the tendency to develop NAT models in real-world systems is increasing.
We will elaborate on these methods in the following sections.

\section{Data Manipulations}
\label{sec:data}
Neural machine translation is a data-driven task, and the performance of the NAT model heavily relies on the volume as well as the quality of the bilingual training data. Therefore, various data manipulation methods are proposed to help the model better capture the target side dependency. 
In this section, we will introduce these methods from two perspectives: (1) knowledge distillation which aims to
reduce the complexity of the training corpus;
(2) data learning strategies that help the model better learn and understand the training data.
The introduced methods are listed in the ``Data Manipulations'' category of Table~\ref{tab:overview}.

\subsection{Knowledge Distillation}
Initially, Knowledge Distillation~(KD)~\cite{hinton2015distilling} is proposed to train a weaker student model with soft prediction distributions generated by a stronger teacher model.
Sequence-level knowledge distillation~\cite{kim2016sequence} extends it to the sentence level, where a pre-trained teacher model predicts sequences of tokens that are taken as the targets of the student model. 
When applied to NAT models, a pre-trained AT model is utilized to generate distilled target sentences for all source sentences in the training set, with either greedy decoding or beam search. 
For example, given a pre-trained AT model $\theta_{\text{AT}}$ and the training set $D=\{(X, Y)_i\}_{i=1}^N$, the distilled target sentences $Y'$ are generated as $Y' \sim P(Y|X;\theta_{\text{AT}}),$
where $Y'$ are decoded with various decoding algorithms such as greedy decoding and beam search decoding. Then, we train the NAR models on this distilled training set $D'=\{(X, Y')_i\}_{i=1}^N$ with the traditional negative log-likelihood loss:
\begin{equation}
  \mathcal{L_{\text{KD}}} = -\sum\nolimits_{(x, y') \in \mathcal{D'}} \log P(y'|x;\theta_{\text{NAT}}), 
\end{equation}
where $\theta_{\text{NAT}}$ is the parameter set of the NAT model.
KD is widely adopted as the distilled corpus is regarded as less noisy and more deterministic than the original one. 
To investigate the reason behind this, we review related works and give a detailed analysis from two aspects: 
(1) why is KD effective for NAT models?
(2) does there exist drawbacks to current KD methods, and how to solve them?

\noindent\textbf{Understanding Knowledge Distillation.}
Zhou~\etal~\cite{zhou2019understanding} 
propose two quantitive measures, including the complexity and faithfulness, to analyze the property of distilled data and its correlation with NAT performance. Specifically, they find that while KD generally simplifies the training data by reducing the complexity and increasing the faithfulness, a larger and better teacher model does not always lead to a better student model. Instead, the capacity of the teacher model should be aligned with the student NAT model to achieve the best performance. 
In addition, Ren~\etal~\cite{ren2020study} design a model to measure the target token dependency over the data and find that KD can reduce the dependency when predicting target tokens, which is helpful for the training of NAT.
Xu~\etal~\cite{xu2021does} find that KD can also reduce the lexical diversity and word reordering degree, which helps the model better learn the alignment between source and target.

\noindent\textbf{Problem and Improvements.}
Despite the effectiveness, there exist some problems when utilizing KD. Zhou~\etal~\cite{zhou2019understanding} find that the capacity of the NAT model should be correlated with the complexity of the distilled dataset.
Therefore they propose several methods, including born-again network~\cite{furlanello2018born} and mixture-of-experts~\cite{shen2019mixture} to adjust the complexity of the dataset w.r.t the model capacity. 
In addition, after knowledge distillation, the density estimation of real data may be harmed~\cite{lee2020discrepancy} and the lexical choice may be mistaken~\cite{ding2020understanding}.
Specifically, Ding~\etal~\cite{ding2020understanding} suppose that the distilled data mainly focuses on the performance of high-frequency words. They propose two evaluation metrics to measure the lexical translation accuracy and conclude that the accuracy of low-frequency words is seriously decreased when the NAT model is trained on distilled datasets.
To deal with this problem, Ding~\etal~\cite{ding2021rejuvenating} make full use of raw, distilled, and reverse-distilled data to rejuvenate low-frequency words. 
Zhou~\etal~\cite{zhou2020improving} add monolingual data for training of the teacher model to enrich the distilled dataset.
The effect and improvement of self-distillation are also explored~\cite{guo2021self}.
Shao~\etal~\cite{shao2022one} also notice that adopting a single distillation reference from a specific teacher is not optimal for training NAT models and thus propose the diverse distillation with reference selection (DDRS) strategy during training.

\begin{table*}[htb]
	\centering
	\renewcommand\arraystretch{1.05}
	\caption{A brief summary and overview of different NAT models discussed in this paper. The numbers of iteration, decoding speedup, and performance are all copied from their original paper. Specifically,  ``Performance'' denotes the BLEU scores on the WMT 14 EN$\to$DE dataset, and ``Speedup'' refers to the decoding speedup ratio compared with AT model. Note that ``*'' indicates training with sequence-level knowledge distillation from a big Transformer. The speedup may not be comparable due to their different hardware conditions, and we list them here just for reference. ``$^\#$'' denotes Findings.}
	\label{tab:overview}
	\vspace{-0.5em}
	\footnotesize
	\resizebox{\linewidth}{!}{
		\begin{tabular}{c|c|c|c|c|ccc}
			\Xhline{1.2pt}
			Category & Sub-category & Method & Description & Publication & Iteration & Speedup & Performance \\
			\hline
			\multirow{11}{*}{\shortstack{Data Manipulations}} & \multirow{5}{*}{Improving KD}  & MD~\cite{zhou2020improving}   & Add monolingual data, enrich distillation corpus   & ACL 2020 & 1 & - & 25.73 \\
			& & RDP~\cite{ding2020understanding} & Raw data prior training, improve lexical choice & ICLR 2020 & 2.5 & 3.5x & 27.80* \\
			& & LRF~\cite{ding2021rejuvenating} & Add reverse-distill data,rejuvenate low-frequency word & ACL 2021 & 2.5 & 3.5x & 28.20* \\
			& & SDMRT~\cite{guo2021self} & Self-distillation mixup, pre-rerank and fine-tune training & ARXIV 2021 & 10 & - & 27.72* \\
                & & DDRS~\cite{shao2022one} & Diverse distillation with reference selection & NAACL 2022 & 1 & 14.7x & 27.60 \\ 
			\cline{2-8}
			& \multirow{6}{*}{Learning Strategies}  & GLAT~\cite{qian2020glancing}   & Glancing, learn from fragments to whole sentence & ACL 2021 & 1 & 15.3x & 25.21 \\
			& & \textit{latent}-GLAT~\cite{bao2022glat} & Introduce glancing strategy to discrete latent variables & ACL 2022 & 1 & 11.3x & 26.64  \\
			& & PMG~\cite{ding2021progressive} & Multi-granularity, from words, phrases, sentences gradually& ACL$^\#$ 2021 & 3.5 & - & 27.80* \\
			& & MvSR-NAT~\cite{xie2022mvsr} & Consistency-based, masked token and model level consistency & TASLP 2022 & 10 & 3.6x & 27.39*\\
            & & LaNMT-C~\cite{zhu2022non} & Consistency training for posterior latent variables & NAACL 2022 & 2 & 11.0x & 26.02 \\
            & & CCMLM~\cite{anonymouscontrastive2022} &  Contrastive common mask and contrastive dropout for CMLM & EMNLP$^\#$ 2022 & 10 & - &27.93* \\
			\hline
			\multirow{30}{*}{\shortstack{Modeling}} & \multirow{11}{*}{\shortstack{Iteration-based\\ methods}}& NAT-IR~\cite{lee2018deterministic}   & Denoising autoencoders, iterative refinement  & EMNLP 2018 & 10 & 1.5x & 21.61 \\
            & & Insertion Transformer~\cite{stern2019insertion}   & Insert tokens each iteration, like balanced binary trees & ICML 2019 & $\approx log_2(N)$ & - & 27.41 \\
			& & CMLM~\cite{ghazvininejad2019mask}  & Masked language model trained with uniform mask strategy & EMNLP 2019 & 10 & 1.7x & 27.03* \\
			& & DisCo~\cite{kasai2020non}  & More visible subsets to predict masked tokens & ICML 2019 & Adaptive & 3.5x & 27.34* \\
            & & SMART~\cite{ghazvininejad2020semi} & Introduce correction task during inference for CMLM & ARXIV 2020 & 10 & - & 27.65 \\
			& & JM-NAT~\cite{guo2020jointly}  & Jointly masked strategy, N-gram level masking in decoder & ACL 2019 & 10 & 5.7x & 27.69* \\
			& & Imputer~\cite{saharia2020non}  & Combine conditional masking with CTC & EMNLP 2020 & 8 & 3.9x & 28.20* \\
			& & R\scriptsize{EWRITE}\footnotesize{NAT}~\cite{geng2021learning}  & Reviewer and locator, locate the error and rewrite & EMNLP 2021 & 2.7 & 3.9x & 27.83* \\
			& & CMLMC~\cite{huang2022improving}  & CMLM with reveal-position and correction function& ICLR 2022 & 10 & - & 28.37* \\
            & & SUNDAE~\cite{savinovstep} & Step-unrolled denoising autoencoder  & ICLR 2022 & 16 & - & 28.46* \\ 
            & & INSNET~\cite{lu2021efficient} & Insertion-oriented relative position encoding, insert in each layer & NeurIPS 2022 & 16.1 & 3.78x & 28.05 \\
			\cline{2-8}
			& \multirow{10}{*}{\shortstack{Latent variable\\-based methods}} & NAT~\cite{gu2018non}   & Fertility predictor, determine the latent alignments & ICLR 2018 & 1 & 15.6x & 17.35  \\
			& & FlowSeq~\cite{ma2019flowseq} & Generative flow, a powerful mathematical framework & EMNLP 2019 & 1  & 1.1x &23.72 \\
			& & PNAT~\cite{bao2019non}   & Positional predictor, model the position of target tokens & ARXIV 2019 & 1 & 7.3x & 23.05* \\
			& & SynST~\cite{akoury2019syntactically} & Parse decoder, autoregressively predict a chunked parse tree & ACL 2019 & $N/6$ & 4.9x & 20.74 \\
			& & LaNMT~\cite{shu2020latent}   & Delta Posterior, continuous latent variables & AAAI 2020 & 1 & 6.8x & 24.20 \\
			& & ReorderNAT~\cite{ran2021guiding} & Reorder the source sentence into a pseudo-translation & AAAI 2021 & 1 & 16.1x & 22.79 \\
			& & AligNART~\cite{song2021alignart} & Aligner module, alignment decomposition strategy & EMNLP 2021 & 1 & 13.4x & 26.40 \\
			& & CNAT~\cite{bao2021non} & Categorical codes, without external syntactic parser & NAACL 2021 & 1 & 10.4x & 25.56* \\
			& & SNAT~\cite{liu2021enriching} & Incorporate the explicit syntactic and semantic structures & EACL 2021 & 1 & 22.6x & 24.64*  \\
			& & Fully NAT~\cite{gu2020fully} & Several tricks to improve the Fully NAT & ACL$^\#$ 2021 & 1 & 16.5x & 27.49*  \\
			\cline{2-8}
			& \multirow{9}{*}{\shortstack{Other enhancements\\-based methods}} & ENAT~\cite{guo2019non} & Phrase-table lookup, embedding mapping  & AAAI 2019 & 1 & 25.3x & 20.65 \\
			& & NAT-REG~\cite{wang2019non}   & Similarity and reconstruction regularization& AAAI 2019 & 1 & 27.6x & 20.65 \\
            & & LAVA NAT~\cite{li2020lava}   & Vocabulary attention, reorder prediction labels of a word & ARXIV 2020 & 1 & 29.3x & 25.72 \\ 
            & & CCAN~\cite{ding2020context}   & Context-aware cross-attention, local and global contexts & COLING 2020 & 10 & - & 27.50 \\
            & & DSLP~\cite{huang2022non}   & Deep supervision, additional layer-wise predictions & AAAI 2022 & 1 & 14.8x & 27.02 \\
            & & DAD~\cite{zhan2022non} & Decoder Input Transformation, backward dependency modeling & ARXIV 2022 &1 & 14.7x & 27.51 \\
            & & DA-Transformer~\cite{huang2022directed} & Represent the hidden states in a directed acyclic graph & ICML 2022 & 1 & 13.9x & 27.49 \\
            & & DA-Transformer Viterbi~\cite{shao2022viterbi} & Adopt viterbi decoding for DA-Transformer & EMNLP$^\#$ 2022 & 1 & 13.2x & 26.89 \\
            & & FA-DAT~\cite{ma2023fuzzy} & Adopt fuzzy alignments for DA-Transformer & ICLR 2023 & 1 & 14.0x & 27.53 \\  
			\hline
			\multirow{9}{*}{Criterion} & \multirow{9}{*}{Loss function} & CTC~\cite{saharia2020non} & Compute and stores partial log-probability & EMNLP 2020 & 1 & 18.7x & 25.60 \\
			&  & BoN~\cite{shao2020minimizing} & N-gram level loss, minimize the Bag-of-Ngrams difference & AAAI 2020 & 1 & 10.8x & 20.90  \\
			&  & AXE~\cite{ghazvininejad2020aligned} & Aligned cross-entropy, a differentiable dynamic program  & ICML 2020 & 1 & 15.3x & 23.53*  \\
			& & EISL~\cite{liu2022don} & Compute the n-gram matching differences, more robust & NAACL 2022 & 1 & - & 24.17* \\
			&  & O\scriptsize{A}\footnotesize{XE}~\cite{du2021order} & Order-agnostic cross-entropy, hungarian algorithm  & ICML 2021 & 1 & 15.3x & 26.10*  \\
            &  & \textit{ngram}-O\scriptsize{A}\footnotesize{XE}~\cite{du2022ngram} & Ngram-based O\scriptsize{A}\footnotesize{XE}, allow reordering between ngram phrases & NAACL 2022 & 1 & 15.2x & 26.50* \\ 
            & & CoCO~\cite{zhang2022study} & Combine CTC and O\scriptsize{A}\footnotesize{XE} loss & NAACL 2022 & 1 & 14.2x & 27.41 \\
            & & MgMO~\cite{li2022multi} & Multi-granularity Metric-based Optimization & EMNLP 2022 & 1 & - & 26.40 \\
            & & NMLA~\cite{shaonon} & Non-monotonic latent alignments, bipartite matching and n-gram matching & NeurIPS 2022 & 1 & 14.7x & 27.57 \\
			\hline
			\multirow{10}{*}{\shortstack{Decoding}} 
			& \multirow{3}{*}{\shortstack{Semi-autoregressive\\ decoding}} & SAT~\cite{wang2018semi}  & Generate muti-tokens at one decoding step & EMNLP 2018 & $N/2$ & 1.5x & 26.90 \\
			& &RecoverSAT~\cite{ran2020learning}& Recover segment, recover mistakes of muti-tokens & ACL 2020 & $N/2$ &2.2x & 27.11\\
            & & GAD++~\cite{xia2022lossless} & Collaboration of NAT drafting and AT verification & ARXIV 2022 & 4.0 & 3.2x & 28.89* \\
			\cline{2-8}
            & \multirow{2}{*}{Insert and delete} 
            &  KERMIT~\cite{chan2019kermit} & Model the joint data distribution, adopt bidirectional fine-tuning & ARXIV 2019 & $\approx log_2(N)$ & - & 28.7 \\
		& & LevT~\cite{gu2019levenshtein}  & Insert and delete tokens during each iteration & NeurIPS 2019 & Adaptive &4.0x & 27.27 \\ \cline{2-8}
            & \multirow{2}{*}{Mask and predict} 
            & Mask-Predict~\cite{ghazvininejad2019mask} & Mask the tokens with low confidence and predict them in the next iteration & EMNLP 2019 & 10 & 1.7x & 27.03* \\
            & & Easy-First~\cite{kasai2020non} & Update tokens at each position with an easy to hard order & ICML 2019 & Adaptive & 3.5x & 27.34* \\ \cline{2-8}
            & \multirow{3}{*}{\shortstack{Mixed decoding}} &  Unified~\cite{tian2020train}  &  Unified approach, conditional permutation language modeling & COLING 2020 & 10 & - & 26.35  \\
            & & Diformer~\cite{wang2022diformer}  & Directional transformer, directional embedding and self-attention  & EAMT 2022 & 10 & - & 27.99  \\
            & & HRT~\cite{wang2022hybrid} & Generate discontinuous sequences autoregressly and fill in others in parallel & ARXIV 2022 & $N/2 + 1$ &  - & 28.49* \\
			\hline
			\multirow{12}{*}{\shortstack{Benefiting from \\ Pre-trained Models }} &\multirow{8}{*}{AT models}  & imitate-NAT~\cite{wei2019imitation}   & Imitation learning framework with imitate module & ACL 2019 & 1  &18.6x & 22.44*  \\
			& &  NAT-HINT~\cite{li2019hint}  & Hints from the hidden state, constrain attention distributions & EMNLP 2019 & 1 & 30.2x & 21.11  \\
			& &  ENGINE~\cite{tu2020engine}  & Energy-based inference, minimize the AT model’s energy & ACL 2020 & - & - & -  \\
			& &  EM+ODD~\cite{sun2020approach}  & Unified framework, dynamically optimize AT and NAT & ICML 2020 & 1 &16.4x & 24.54  \\
            & &  FCL-NAT~\cite{guo2020fine} & Curriculum learning from better-trained state of AT model & AAAI 2020 & 1 &28.9x & 21.70  \\
			& &  MULTI-TASK NAT~\cite{hao2021multi}  & Shared encoder, dynamically mix two training loss   & NAACL 2021 &10 &- & 27.98*  \\
			& &  TCT-NAT~\cite{liu2020task}  & Task-level curriculum learning, from AT to SAT, then to NAT  & IJCAI 2021 & 1 &27.6x & 21.94 \\ 
            & & weak MTL~\cite{wang2022helping} & Multitask learning framework, provide more informative learning signals &EMNLP 2022 & 1 & - & 27.25 \\
			\cline{2-8}
			& \multirow{4}{*}{\shortstack{Pre-trained language\\ models}}  & AB-Net~\cite{guo2020incorporating}  & Take two different BERT models as the encoder and decoder & NeurIPS 2020 & - & 2.4x & 28.69*  \\
			& & NAG-BERT~\cite{su2021non} & Employ bert as a backbone, add a CRF Layer & EACL 2021 & - & - & - \\
			& & CeMAT~\cite{li2022universal} & Aligned code-switching and masking, dynamic dual-masking & ACL 2022 & 10 & - & 27.20 \\
                & & XLM-D~\cite{wang2022xlm} & Lightweight yet effective decorator, adapt the XMLR model into NAT models & EMNLP 2022 &  8 & 2.8x & 29.80 \\
			\hline

		\end{tabular}
	}
\vspace{-1.0em}
\end{table*}
\subsection{Data Learning Strategies}
\label{sec:datalearn}
Aside from constructing informative training datasets, designing suitable learning strategies is another way to improve NAT models. We introduce various data learning strategies 
in this subsection.
Curriculum learning~\cite{bengio2009curriculum} is a machine learning strategy inspired by human learning, which trains the model by feeding training instances in an order (e.g., from easy to hard) instead of randomly.
Guo~\etal~\cite{guo2020fine} introduce the idea of curriculum learning into the training of NAT models by progressively switching the decoder input from AT to NAT to provide a smooth transformation between two training strategies. Liu~\etal~\cite{liu2020task} extend this method by designing more fine-grained curriculums.
Qian~\etal~\cite{qian2020glancing} propose an adaptive glancing sampling strategy to guide the model to learn from fragments first and then from whole sentences gradually.
The ratio of fragments is correlated with the capacity of the model at the current training stage.
Bao~\etal~\cite{bao2022glat} further extend this glancing sampling strategy to a variable-based model.
Song~\etal~\cite{song2022switchglat} combine this glancing sampling strategy with a code-switch method for the task of multilingual machine translation.
Ding~\etal~\cite{ding2021progressive} divide training data into multiple granularities, such as words, phrases, and sentences, and propose a progressive multi-granularity training strategy to train the model from easy to hard.
Apart from curriculum learning,
consistency training is an effective method for autoregressive NMT models~\cite{wu2021r}.
For NAT models, Xie~\etal~\cite{xie2022mvsr} utilize consistency training to improve the
training consistency on different masked sentences. They assumed that the prediction of the same masked position should be consistent in different contexts or with different models. 
A similar idea is also explored for variational autoencoder-based latent-variable NAT models in recent papers~\cite{zhu2022non}, which propose posterior consistency regularization to improve the ability of models. They first apply data augmentation on both source and target sentences twice and then predict the latent variable and regularize these two results.
Besides, contrastive learning is also adopted to improve the performance of NAT models~\cite{anonymouscontrastive2022}, which optimizes the similarity of several different representations of the same token in the same sentence, resulting in more informative and robust representations.

\section{Modeling}
\label{sec:model}
Model structure plays a critical role for NAT models to better capture the target side dependency. This section first introduces two popular frameworks for NAT: iteration-based methods and latent variable-based methods, then we summarize the efforts made on other enhancements-based methods for NAT models. 
The introduced methods are listed in the ``Modeling'' category of Table~\ref{tab:overview}.
Representative methods are illustrated in Figure~\ref{fig:modeling} of the Appendix.

\subsection{Iteration-Based Methods}
\label{sec:iteration}
Iteration-based methods aim to find the trade-off between translation speed and quality.
Instead of generating all target tokens in one pass, they
learn the conditional distribution over partially observed generated tokens. 
Lee~\etal~\cite{lee2018deterministic} first propose the iterative model, they utilize either the output of the previous iteration or the noised target sentence to initial the decoder input for refinements. Besides, iteration-based models can be divided into the following categories:

\noindent\textbf{Text Editing}.
\label{sec:insermodel}
These methods learn to generate tokens with different atomic operations.
Stern~\etal~\cite{stern2019insertion} propose Insertion Transformer which models both what to insert and where to insert relative to the current slot representations via concatenated outputs. They also introduce several order loss functions for training.
Chan~\cite{chan2019kermit} introduce KERMIT, which is similar to Insertion Transformer but models the joint data distribution and its decompositions.
Welleck~\etal~\cite{welleck2019non} frame the insertion learning problem as an imitation learning problem, in which a generation policy is learned to mimic the actions of an oracle generation policy. They propose an annealed coaching method and a roll-in and roll-out procedure to learn insertion. 
Besides, Gu~\etal~\cite{gu2019insertion} propose an insertion-based model with Inferred Generation Order (InDIGO), where the generation orders are modeled as latent variables. InDIGO can automatically infer the generation orders by simultaneously predicting a word and its position to be inserted during inference.
Deletion operation is introduced in \cite{gu2019levenshtein}. They propose Levenshtein Transformer (LevT), which adopts dual policy learning for training and three different classifiers to decide where and how many tokens to insert, whether to delete the tokens and predict the tokens. 
Furthermore, much progress has been made in exploring more potential of the Levenshtein Transformer. 
Xu~\etal~\cite{xu2022bilingual} view translation as a bilingual synchronization task and propose Edit-LevT to explore the potential in a non-autoregressive manner, which adopts Levenshtein Transformer as a backbone model.
Later they also propose TM-LevT~\cite{xu2022non}, 
where Levenshtein Transformer with an additional initial deletion operation is adopted to detect potential irrelevant words in the translation memory.
Niwa~\etal~\cite{niwa2022near} adopt the nearest neighbor as the initial state of the NAR decoder. They introduce NeighborEdit, which retrieves the nearest neighbor of an input sentence and edits it to generate the output sentence.
Lu~\etal~\cite{lu2021efficient} propose a more efficient, flexible, and performance insertion-based model, which introduces an insertion-oriented position encoding method and a better algorithm to determine the parallelization of insertion operations.

\noindent \textbf{Masked Language Modeling.}
Another line of work leverages the success of masked language modeling, initially proposed by BERT~\cite{kenton2019bert}.
Ghazvininejad~\etal~\cite{ghazvininejad2019mask} extend it to the conditional masked language model (CMLM) by masking and predicting target tokens during training. Unlike the fixed masking ratio in BERT, CMLM adopts a uniform masking strategy to capture the interdependencies of target tokens during training. 
Based on this model, several follow-up works are proposed, including: (1) jointly masking tokens~\cite{guo2020jointly}, where the tokens in the source sentences are also masked; 
(2) introducing self-review mechanism~\cite{xie2020infusing}, which applies an AR-decoder to help infuse sequential information;
(3) predicting more visible subsets~\cite{kasai2020non}, instead of only predicting the masked tokens, the method predicts every target token;
(4) introducing self-correction task~\cite{huang2022improving}, rather than only predicting the masked tokens, which can learn to correct the unreasonable tokens generated by inputting a fully masked sequence.

\subsection{Latent Variable-Based Methods.}
\label{sec:latent}
Utilizing latent variables as part of the model is also a popular method to reduce the target side dependency.
Latent variable models maximize the following likelihood:
\begin{equation}
    \mathcal{L}_{\text{Lat}}=\sum\nolimits_{t=1}^{T} \log p(Z|X;\theta) p(y_t|Z,X;\theta),
\end{equation}
where $Z$ is a specific latent variable. 
The latent variable-based NAT models first predict a latent variable sequence, where each variable may be a chunk of words or include some other prompt information.
Existing works mainly apply latent variables to capture the following information.

\noindent\textbf{Prior Target Linguistic Information.}
Ma~\etal~\cite{ma2019flowseq} utilize a powerful mathematical framework called generative flow. 
Variational auto-encoders (VAE) based methods are also applied to model the dependency~\cite{kaiser2018fast}. 
Shu~\etal~\cite{shu2020latent} and Lee~\etal~\cite{lee2020discrepancy} model the latent variables as spherical Gaussian for every token in the encoder.
Bao~\etal~\cite{bao2022glat} utilize a glancing sampling strategy to optimize latent variables. 

\noindent\textbf{Alignments between Source and Target Sentences.}
Gu~\etal~\cite{gu2018non} pre-define the latent variable $Z$ as fertility and use it to determine how many target words every source word is aligned to. Song~\etal~\cite{song2021alignart} predict the alignment by an aligner module as the latent variable $Z$. 

\noindent\textbf{Position Information of Target Tokens.}
Bao~\etal~\cite{bao2019non} propose PNAT, which depends on the part of an extra positional predictor module to achieve the permutation $Z$. 
Ran~\etal~\cite{ran2021guiding} propose ReorderNAT, a novel NAT framework that reorders the source sentence by the target word order to help the decision of word positions.

\noindent\textbf{Syntactic Information of Target Sentence.}
Syntactic labels represent the sentence structure, which can be utilized to guide the generation and arrangement of target tokens.
Akoury~\etal~\cite{akoury2019syntactically} first introduce syntactic labels as a supervision to help the learning of discrete latent variables. 
However, the method needs an external syntactic parser to produce the syntactic reference tree, which is effective only in limited scenarios. To release the limitation, 
Bao~\etal~\cite{bao2021non} propose to learn a set of latent codes that act like the syntactic label.
Liu~\etal~\cite{liu2021enriching} incorporate the explicit syntactic and semantic structures to improve the ability of NAT models. Specifically, they utilize Part of Speech~(POS) and Named Entity Recognition~(NER) to introduce these information.

\subsection{Other Enhancements-based Methods}
\label{sec:enhancement}
In addition to the above two popular frameworks for NAT models,
many efforts have been made to improve the ability of capturing the target side dependency for NAT models at different stages, and the corresponding module is also added to their models. 
We summarize these methods into the following categories.

\noindent\textbf{Enhancing the Input of Decoder.}
Since copying the source sentence to initial the decoder cannot offer any target information~\cite{gu2018non}, Guo~\etal~\cite{guo2019non} propose phrase-table lookup and embedding mapping methods to enhance the input of the decoder, which can feed tokens with some target information into the decoder, then help model learn the training data better. While the used phrase table is trained in advance, embedding mapping drew lessons on the idea of adversarial training and can perform word-level constraints to close the input and target sentence.
Zhan~\etal~\cite{zhan2022non} also focus on the input of the decoder. They propose decoder input transformation, which transforms the decoder input into the target space. Then this can close the input and target side embedding and help capture the target side dependency.

\noindent\textbf{Supervising the Intermediate States.}
Several works give extra guidance to the decoder module. 
Firstly, additional attention modules are applied to learn more information.  
Li~\etal~\cite{li2020lava} propose the Vocabulary Attention (VA) mechanism along with the Look-Around (LA) strategy to help the model capture long-term token dependencies of the target sentence.
Ding~\etal~\cite{ding2020context} propose a context-aware cross-attention module that focuses on both local and global contexts simultaneously and therefore enhances the supervision signal of neighbor tokens as well as the information provided by the source texts.
Besides, Huang~\etal~\cite{huang2022non} provide layer-wise supervision to the intermediate states of each decoder layer.

\noindent\textbf{Improving the Output of Decoder.}
For the output of the decoder, Wang~\etal~\cite{wang2019non} regularize the learning of the decoder representations by introducing similarity and reconstruction regularizations, where the former aims at avoiding similar hidden states to alleviate the repetitive translation problem, and the latter constraints the results to help address the problem of incomplete translations.
Besides, Ran~\etal~\cite{ran2020learning} propose the RecoverSAT model to recover from repetitive and missing token errors by dynamically determining the length of segments that need to recover and then deleting repetitive segments.
Huang~\cite{huang2022directed} propose Directed Acyclic TransfoRmer (DA-Transformer), which represents the hidden states in a Directed Acyclic Graph (DAG), and each path of the DAG denotes a specific translation. This method dramatically helps capture the dependency of target tokens. Recently, more enhancing methods based on DA-Transformer have also been proposed~\cite{shao2022viterbi,ma2023fuzzy}.
Shao~\etal~\cite{shao2022rephrasing} introduce a rephraser to provide a better training target for NAT models. They apply reinforcement learning to obtain a good rephraser and then train NAT models based on the rephraser output.

\section{Criterion}
\label{sec:criterion}
In addition to training data and model structure, training criterion is always another decisive factor for the success of neural network models. 
Most NMT models apply cross-entropy (CE) loss as their training criterion:
\begin{equation}
    \mathcal{L}_{\text{CE}}= - \sum\nolimits_{t=1}^{T} \log P(y_t|X;\theta),
\end{equation}
where each $P(y_t|X;\theta)$ is calculated conditional independently by
the NAT model with parameters $\theta$.
However, several researchers have pointed out that the traditional CE loss may not be optimal for NAT models 
and they propose better criteria to improve the performance of NAT models. 
This section compares these criteria with traditional CE loss, emphasizes their advantages, and summarizes them into the following categories.

\noindent\textbf{Connectionist Temporal Classification~(CTC).}
CTC based criteria~\cite{graves2006connectionist}
compute and store partial log-probability summations for all prefixes and suffixes of the output sequence by dynamic programming to alleviate the misalignment problem.
Libovicky~\etal~\cite{libovicky2018end} and Shu~\etal~\cite{shu2020latent} also use CTC loss to marginalize all the monotonic alignments between target and predictions,
which can be written as 
\begin{equation}
    \mathcal{L}_{\text{CTC}}= - \sum\nolimits_{a\in \beta(y)}\prod_{i} p(a_{i}|x,\theta)),
\end{equation}
where $a$ is a possible latent alignment, $\beta(y)$ denotes all possible alignments based on the CTC format.
Shao~\etal~\cite{shaonon} further explore non-monotonic latent alignments and propose two matching objectives named bipartite matching and n-gram matching to enhance the training of NAT models.

\noindent\textbf{N-Gram-Based.}
N-gram-based criteria~\cite{shao2020minimizing} focus on n-gram level relationships. The word-level CE loss encourages NAT to generate the target tokens without considering the global correctness, which aggravates the weakness in capturing target side dependency.
Shao~\etal~\cite{shao2020minimizing} propose an n-gram level loss function to minimize the Bag-of-Ngrams (BoN) difference between the model output and the reference sentence. Guo~\etal~\cite{guo2020jointly} also introduce the n-gram-based dependency of target tokens to alleviate the problem of repetitive translations.
N-gram-based loss can be written as: 
\begin{equation}
  \mathcal{L}_{\text{BoN}}= \dfrac{\text{BoN-}L1}{2(T - n + 1)},  
\end{equation}
where $\text{BoN-}L1$ is the L1 distance between the number of n-grams predicted by the NAT model and that in the reference sentence, which can be calculated as:
\begin{equation}
    \text{BoN-}L1 = \sum\nolimits_{g} |\text{BoN}_{\theta}(g) - \text{BoN}_{Y}(g)|,
\end{equation}
where $g= (g_1, g_2, ... , g_n) $ is a possible n-gram set. $\text{BoN}_{Y}(g)= \sum_{t=0}^{T-n} 1\{y_{t+1:t+n} = g\}$ is the number of occurrences of $g$ in sentence $Y$. $\text{BoN}_{\theta}(g)$ denotes the BoN for a NAT model with parameters $\alpha$, which can be written as:
\begin{equation}
    \begin{aligned}
       \text{BoN}_{\theta}(g) &= \sum\nolimits_{Y} P(Y|X,\theta)*\text{BoN}_{Y}(g) \\
       &= \sum\nolimits_{Y} P(Y|X,\theta) * \sum\nolimits_{t=0}^{T-n} 1\{y_{t+1:t+n} = g\} \\
       &= \sum\nolimits_{t=0}^{T-n} \prod\nolimits_{i=1}^{n} P(y_{t+i = g_i|X,\theta})
    \end{aligned}
\end{equation}
where $X$ and $Y$ denote the source and target sentences, respectively.
Liu~\etal~\cite{liu2022don} propose a novel Edit-Invariant Sequence Loss (EISL)
which focuses on the n-gram matching to make the model 
perform more robustly when encountering inconsistent sequence order of source and target.
They show that NAT benefits from this loss since the vanilla NAT model is struggling to model flexible generation order.

\noindent\textbf{Order-Based.}
CE loss is sensitive to any inconsistent alignments between the prediction and target, which leads to penalizing a reasonable translation if it only mismatches the positions of target tokens. 
To soften the penalty for word order errors, Ghazvininejad~\etal~\cite{ghazvininejad2020aligned} propose aligned cross-entropy (AXE) loss, which uses a differentiable dynamic programming method to determine loss based on the best possible monotonic alignment between the ground-truth and the model predictions. 
The AXE loss is calculated as:
\begin{equation}
\mathcal{L}_{\text{AXE}}= - \sum\nolimits_{t=1}^{T} \log P_{\alpha}(y_t|X;\theta) - \sum\nolimits_{k\notin\theta}  P_{k}(\epsilon),
\end{equation}
where the first term indicates the aligned cross-entropy loss function between the target tokens and predictions, and the second term penalizes the unaligned predictions.
Besides, Du~\etal~\cite{du2021order} further propose the order-agnostic cross-entropy (O\scriptsize{A}\normalsize XE) loss, which applies the Hungarian algorithm to find the best possible alignment. The O\scriptsize{A}\normalsize XE loss almost removes the penalty for order errors and guides NAT models to focus on lexical matching. 
Du~\etal~\cite{du2022ngram} also extend O\scriptsize{A}\normalsize XE loss by allowing reordering between n-gram phrases but maintaining a strict match of word order within phrases.
Li~\etal~\cite{li2022multi} additionally learn to alleviate the constraint of strict match between the hypothesis
and the reference tokens. They propose Multi-granularity Metric-based Optimization (MgMO) to collect model behaviors on varied granularity of translation segments and use the feedback for back-propagation.

Given a parallel training sample $(X, Y)$, we can define the alignment between a model prediction $\hat{Y} = \{ \hat{y}_1, \hat{y}_2,..., \hat{y}_{T_{\hat{Y}}}\}$ and a target sentence $Y=\{y_1, y_2, ..., y_{T_Y}\}$ as an ordering of the set of target tokens $Y$, e.g., $O^i = \{y_{T_Y}, y_1,..., y_{T_{Y-1}}\}$ denotes that tokens $\hat{y}_1, \hat{y}_2,..., \hat{y}_{T_{\hat{Y}}}$ in model prediction $\hat{Y}$ are aligned with tokens $y_{T_Y}, y_1,..., y_{T_{Y-1}}$ in target sentence $Y$ respectively. Note that during training, $T_Y = T_{\hat{Y}}$. For each target sentence, we can get $T_Y!$ monotonic alignments. Based on each alignment state $O^i$, the corresponding CE loss can be calculated as $\mathcal{L}_{\text{O}^i}= - \log P(O^i|X;\theta)$.
Given all possible alignment states $O = \{O^1, O^2,...,O^{T_Y!}\}$, 
the O\scriptsize{A}\normalsize XE objective is defined as finding the best alignment $O^i$ to minimize:
\begin{equation}
    \mathcal{L}_{\text{O\tiny{A}\scriptsize XE}}= \mathop{\arg\min}\limits_{O^i \in O}(\mathcal{L}_{O^i})
\end{equation}
where $- \log P(O^i|X;\theta)$ indicates the CE loss with ordering $O^i$.
The above methods are listed in the ``Criterion'' category of Table~\ref{tab:overview} and exemplified in Figure~\ref{fig:loss} of the Appendix.

\section{Decoding}
\label{sec:decoding}
The decoding stage is also crucial for neural machine translation models.
Some works try to improve the NAT decoding schedule by applying different tricks. 
As mentioned in section~\ref{sec:comparison}, NAT models need to know the target length to guide decoding. And after the length is predicted, different decoding schedules are adopted to improve decoding. In this section, we will introduce various length prediction methods and decoding strategies.

\subsection{Length Prediction}
In AT models, the beginning and end of decoding are controlled by special tokens, including \texttt{[BOS]} (beginning of a sentence) and \texttt{[EOS]} (end of a sentence), which implicitly determine the target length during decoding.
However, as all target tokens are generated in parallel in NAT models, there is no such special token or target information to guide the termination of decoding. 
NAT models must know the target length in advance and then generate the content based on it.   
Therefore, how to predict the correct length of the target sentence is critical for NAT models~\cite{wang2021length}. Different methods for target length prediction have been proposed.

\noindent\textbf{Length Prediction Mechanism.}
\label{sec:length_prediction}
Length information of target sentence is essential to NAT models as mentioned above.
Gu~\etal~\cite{gu2018non} propose a fertility predictor to decide how many times the source token will be copied when constructing the decoder input. Then, the sum of fertility numbers could be viewed as the length of the target sentence. Other length prediction methods are also proposed:
(1) Classification modeling, which formulates the length prediction as a classification task and utilizes the encoder output to predict the target length or the length difference between the source and target ~\cite{lee2018deterministic,wei2019imitation};
(2) Linear modeling, Sun~\etal~\cite{sun2019fast} try to use a linear function such as $T_y =\alpha~T_x + B$ to directly calculate the target length based on source length;
(3) Special token modeling by introducing a special \texttt{[LENGTH]} token~\cite{ghazvininejad2019mask,tu2020engine,qian2020glancing}. Akin to the \texttt{[CLS]} token in BERT, the \texttt{[LENGTH]} token is usually appended to the encoder input, and
the model is trained to predict the length of the target sentence utilizing the hidden output of the \texttt{[LENGTH]} token; 
(4) CTC-based modeling, several models implicitly determine the target length from the word alignment information~\cite{libovicky2018end,saharia2020non} based on the connectionist temporal classification (CTC)~\cite{graves2006connectionist} results.

\noindent\textbf{Length Prediction Improvements.}
Inevitably, there is a deviation between the predicted length and the true length. 
To release the inherent uncertainty of the data itself, length parallel decoding (LPD)~\cite{lee2018deterministic,guo2019non} and noise parallel decoding (NPD)~\cite{gu2018non,ghazvininejad2019mask} are widely utilized during inference. 
(1) LPD is often used in classification-based models. 
Once the length $T$ is determined, they choose an LPD window $m$ and then obtain multiple translation results with lengths in the range ${[T-m,T+m]}$. A pre-trained autoregressive model is then used to score and select the best overall translation.
(2) Models that adopt NPD choose the top $m$ lengths with the highest length prediction probability and return the translation candidate with the highest log probabilities on the average of all tokens.

\subsection{Decoding Strategy}
Fully NAT models adopt only one-step decoding, which can greatly speed up decoding but fail to achieve high-quality translation. As shown in Table \ref{tab:overview}, the performance of iteration-based models is generally better than that of fully NAT models, indicating that NAT models fail to capture the target side dependency correctly with only one-step decoding.

\noindent\textbf{Semi-Autoregressive Decoding.}
Semi-autoregressive decoding is adopted for SAT models, which generates multiple target tokens at one decoding step. This decoding manner does not remove the dependency of target tokens completely. Several methods are proposed based on the semi-autoregressive decoding manner, such as: (1) Syntactic labels based~\cite{akoury2019syntactically},
which applies a syntactic parser to produce
the syntactic reference tree for the tokens in the current decoding step, then a group of tokens with a close syntactic relationship will be generated at one step. (2) Recover mechanism~\cite{ran2020learning}, which aims to alleviate the multi-modality problem by introducing a recovered segment. Once a group of tokens is generated, the model will recover from
missing and repetitive token errors. 
(3) Aggressive decoding~\cite{xia2022lossless}, which first aggressively decodes several tokens as a draft in a non-autoregressive manner and then verifies them in an autoregressive manner. This method can improve the translation quality and lower the latency as the drafting and verification can execute in parallel.

\noindent\textbf{Insert and Delete.}
Insertion-based decoding methods aim to insert tokens during each decoding step. 
Many works explore this method based on different generation orders, including uniform~\cite{welleck2019non}, random~\cite{gu2019insertion}, or balanced binary trees~\cite{stern2019insertion,chan2019kermit}. More exploration of insertion orders can be found in ~\cite{chan2020empirical}.
Besides, several advanced methods are proposed:
(1) introducing deletion operations~\cite{gu2019levenshtein}, which also allows the model to delete the
unreasonable tokens during each decoding step;
(2) adaptive parallelization of insertions~\cite{lu2021efficient}, where parallel insertion is conducted between each decoder layer to better improve decoding efficiency.

\noindent\textbf{Mask and Predict.}  
Ghazvininejad~\etal~\cite{ghazvininejad2019mask} first propose a mask-predict algorithm. Starting from a fully masked sequence, the model aims to predict the masked tokens during each iteration. Then a fraction of target tokens with low prediction probability will be masked again and fed to the decoder for the next iteration. Many related works try to improve the performance of this decoding method:
(1) easy-first policy~\cite{kasai2020non}, which modifies the mask prediction algorithm by updating each position with an easy to hard order given the prediction probability of previous iterations;
(2) correcting the unmasked tokens~\cite{ghazvininejad2020semi,huang2022improving}, where the unmasked tokens are self-corrected during each iteration;
(3) substitutive masking strategies~\cite{kreutzer2020inference}, where several strategies are proposed to explore the effect of numbers and rules of masking tokens;
(4) utilizing the locator module~\cite{geng2021learning}, which also focuses on the importance of determining the tokens replaced by \texttt{[mask]} tokens in the next iteration and transforms it into a binary classification problem.
Inspired by the beam search algorithm for the CTC-based model, Kasner~\etal~\cite{kasner2020improving} apply beam search and employ additional features in its scoring model to improve the fluency of NAT.
 
\noindent\textbf{Mixed Decoding.} 
Since different types of decoding strategies have been proposed for NAT, several works aim to combine these decoding strategies into a unified model~\cite{tian2020train,wang2022diformer}.
Tian~\etal~\cite{tian2020train} propose a unified approach for machine translation that supports autoregressive, semi-autoregressive, and iterative decoding methods. Once the model is trained, any of the above decoding strategies can be applied by repeatedly determining positions and generating tokens on them. 
Taking a step further, Wang~\etal~\cite{wang2022diformer} propose a directional Transformer, which models the AR and NAR generation with a unified framework by designing a special attention module. Their model supports four decoding strategies and can dynamically select strategies during each iteration.
Wang~\etal~\cite{wang2022hybrid} combine the strengths of autoregressive and non-autoregressive translation paradigms well. They propose 
hybridregressive translation method, which first generates discontinuous sequences autoregressively and then fills in all previously skipped tokens in parallel.
We depict typical decoding strategies in Figure~\ref{fig:decoding} of the Appendix and use an example to show their differences in the Appendix Figure~\ref{fig:example}.

\section{Benefiting from Pre-trained Models}
\label{sec:benefition}
To improve the performance of NAT models, various methods are proposed to leverage the information from other strong models, such as their AT counterparts and large-scale pre-trained language models. We will introduce these methods in the following content.
\subsection{AT Models}
\label{sec:at}
Due to the strong performance of AT models, leveraging AT models to help the NAT model training is appealing. 
But they differ in model structure and decoding strategy. Therefore, different techniques to benefit the NAT model from their AT counterparts are proposed:

\noindent\textbf{Training with the Supervision of AT Models.}
Wei~\etal~\cite{wei2019imitation} propose a novel imitation learning framework, introducing a better trained AT demonstrator to supervise each decoding state of the NAT model across different times so that the problem of huge search space can be alleviated. 
Li~\etal~\cite{li2019hint} design two kinds of hints from the hidden representation level to regularize the KL-divergence of the encoder-decoder attention between the AT and NAT models, which can help the training of NAT models. 
Besides, Tu~\etal~\cite{tu2020engine} propose an energy-based inference network to minimize the energy of AT model and give several methods for relaxing the energy.

\noindent\textbf{Fine-Tuning from AT Models.}
Guo~\etal~\cite{guo2020fine} utilize curriculum learning to fine-tune from a better-trained state of AT models, and two curricula for the decoder input and mask attention are applied.
In addition, Liu~\etal~\cite{liu2020task} propose task-level curriculum learning to shift the training strategy from AT to SAT gradually, and finally to NAT.

\noindent\textbf{Training with AT Models Together.}
Sun~\etal~\cite{sun2020approach} propose a unified Expectation-Maximization (EM) framework. It optimizes both AT and NAT models jointly, which iteratively updates the AT model based on the output of the NAT model and trains the NAT model with the new output of AT model. Besides, Hao~\etal~\cite{hao2021multi} propose a model with a shared encoder and separated decoders for AT and NAT models. The training for these two models is controlled by different weights to mix two training losses. 
A similar idea is also adopted in \cite{wang2022helping}, but they introduce a weak AR decoder module that predicts the tokens entirely depending on the information provided by the NAR decoder to improve the capability of the NAR decoder.

\subsection{Pre-Trained Language Models}
While large-scale pre-trained language models have been proven effective in autoregressive machine translation~\cite{zhuincorporating,yang2020csp}, efforts are also made for non-autoregressive machine translation. Guo~\etal~\cite{guo2020incorporating} incorporate BERT into machine translation based on the mask-predict decoding method, which initializes the encoder and decoder with corresponding pre-trained BERT models, and inserts adapter layers into each layer.
Su~\etal~\cite{su2021non} employ BERT as the backbone model and add a CRF output layer for better capturing the target side dependency to improve the performance further.
Li~\etal~\cite{li2022universal} propose a conditional masked language model with an aligned code-switching masking strategy to enhance the cross-lingual ability. The proposed model can be fine-tuned on both NAT and AT tasks with promising performance.
Wang~\etal~\cite{wang2022xlm} present XLM-D, where a lightweight yet effective decorator is adopted to adapt the cross-lingual pretraining model (XLMR) into NAT models.

\section{Summary of Non-Autoregressive NMT}
\label{sec:summary}
All of the above techniques can mitigate the challenge of failing to capture the target side dependency more or less by reducing the reliance of NAT models on target tokens.
Since NAT models are essentially data-driven, their performance highly depends on data volume, quality, and learning strategies.
Thus, data manipulation methods are almost indispensable for existing NAT works. 
Various KD methods can reduce the complexity of the training corpus, while data learning strategies can facilitate the understanding and learning of training data.
Another critical element in capturing the target side dependency is NAT model structure, e.g., iteration-based methods, latent variables, and various add-ons for the decoder module.
In addition to data manipulation and model structure, better training criteria are proposed to make up for the deficiency of cross-entropy loss, e.g., leveraging CTC-based criteria to alleviate the misalignment problem, introducing n-gram-based criteria to capture global context other than word-level correctness, and designing order-based criteria to soften the penalty for reasonable translations but with mismatched tokens at the target positions.
Since the differences between AT and NAT models are mainly manifested in the decoder part, different improving skills for the NAT decoding mode are also presented.
Typical strategies include performing target length prediction to guide the end of decoding and improving the one-step decoding by keeping part of the target side dependency information in semi-autoregressive decoding, providing partial target information in iterative decoding, and exploring their combinations in mixed decoding.
Besides, leveraging the information from other strong models can further improve the performance of NAT models, such as utilizing information from their AT counterparts and large-scale pre-trained language models. 
To help researchers and engineers select appropriate techniques in applications, we also conduct a brief comparison between existing methods on their effectiveness and inference speed in Appendix based on the contents of Figure~\ref{fig:bleuvsspeed} and Figure~\ref{fig:frmework}.

Recent research has also disclosed the potential problems of NAT models.
One recent concern is that the decoding speedup ratio mentioned in Table~\ref{tab:overview} is measured by $L_1^{\text{GPU}}$, i.e., the translation latency of one input sentence by running the model on a single GPU.
With the increase of batch size, the superiority of NAT models against AT alternatives on inference efficiency will be shortened~\cite{helcl2022non}.
The speedup reported on different hardware architectures also makes the comparison less persuasive.
Moreover, the inference speedup of NAT models is generally measured on AT frameworks with a symmetrical encoder and decoder structure with the same number of transformer layers without considering these more efficient AR models with shallow decoders \cite{kasai2020deep}.  
Besides the lack of evaluation fairness in latency, other defects in translation quality evaluation also exist. 
Helcl~\etal~\cite{helcl2022non} point out that the NAT models produce unusual errors that the widely used  BLEU~\cite{papineni2002bleu} metric does not penalize very heavily, leading to spurious comparability in translation quality between AT and NAT models.
It is necessary to introduce more feasible metrics~\cite{helcl2022non,schmidt2022non} such as 
CHrF~\cite{popovic2015chrf}, ScareBLEU~\cite{post2018call}, COMET~\cite{rei2020comet} to obtain a reliable and comprehensive evaluation.
Last but not least, fine-grained evaluation of translation quality from different aspects, e.g., word repetition, sentence fluency, lexical choice, sentence consistency, and syntactic accuracy, are underexplored but critical to move NAT forward. 
\section{Extensive Applications Beyond NMT}
\label{sec:extension}
After seeing the success of non-autoregressive (NAR) techniques on neural machine translation, these strategies are also widely applied to extensive text generation tasks, semantic parsing~\cite{babu2021non,shrivastava2021span}, text to speech~\cite{ren2019fastspeech,peng2020non}, etc. 
In this section, we will conduct a brief discussion about these works.

\subsection{Text Generation}
The inference efficiency is not only required for neural machine translation but also indispensable for many other text generation tasks~\cite{yang2021pos,jiang2021improving,su2021non}. 
Existing works of introducing NAR techniques into text generation tasks focus on automatic speech recognition~\cite{higuchi2021comparative,yu2021non,song2021non}, text summarization~\cite{qi2021bang}, grammatical error correction~\cite{straka2021character,li2021tail}, dialogue~\cite{han2020non,zou2021thinking}.
Resembling the encountered challenge of NAT models in Section \ref{sec:main_challenge}, representative works of non-autoregressive text generation mainly address the problems of missing target side information and length prediction.
According to the involved tasks, we structure these works into different groups, including general-purpose NAR methods and typical models for each specific generation task.

\noindent\textbf{General-Purpose NAR Text Generation.}
Some works aim to design a general NAR method that can support multiple text generation tasks. 
Su~\etal~\cite{su2021non} employ BERT as the backbone of a NAR generation model for machine translation, sentence compression, and summarization.
They add a CRF output layer on the BERT architecture for non-autoregressive tasks. For length prediction, they adopt two special tokens \texttt{[eos]} to dynamically guide the end of the generation.
They extend the architecture of BERT to capture the target side dependency better and improve the performance further.
For length prediction, they propose a simple and elegant decoding mechanism to help the model determine the target length on-the-fly.
Jiang~\etal~\cite{jiang2021improving} propose a new paradigm to adopt pre-trained encoders for NAR text generation tasks. 
They propose a simple and effective iterative training method, MIx Source and pseudo Target (MIST), for the training stage without introducing extra cost during inference. 
Yang~\etal~\cite{yang2021pos} attempt to explore the alternatives for KD in text summarization and story generation. They focus on linguistic structure predicted by a Part-of-Speech (POS) predictor to help 
alleviate the multimodality problem.
Mallinson~\etal~\cite{mallinson2022edit5} propose EdiT5, which decomposes the generation process into three sub-tasks: tagging, re-ordering, and insertion, where the first two adopt a non-autoregressive manner, but the last one uses an autoregressive decoder. 
They evaluate the performance in sentence fusion, grammatical error correction, and decontextualization.
Qi~\etal~\cite{qi2021bang} explore to design a large-scale pre-trained model that can support different decoding strategies when applied to downstream tasks. Concretely, they leverage different attention mechanisms during the training stage and fine-tuning strategies to adapt from AR to NAR generation. To verify the effectiveness of their model, they evaluate the proposed method for question generation, summarization, and dialogue generation tasks. 
Moreover, they further introduce a self-paced mixed distillation method~\cite{qi2022self} to improve the generation capability of BANG.
Agrawal~\etal~\cite{agrawal2022imitation} propose a framework that adopts an imitation learning algorithm for applying NAR models to editing tasks such as controllable text simplification and abstractive summarization. 
They introduce a roll-in policy and a controllable curriculum to alleviate the mismatching problem between training and inference.
Li~\etal~\cite{li2022elmer} propose ELMER, which introduces a token-level early exit mechanism into NAR models for the first time. They leverage layer permutation language modeling for pre-training, which can achieve substantial performance improvements on text summarization, question generation, and dialogue generation tasks.

\noindent\textbf{Task-Specific NAR Text Generation.}
\label{sec:Task-Specific}
Many other works introduce NAR methods for a specific text generation task.
\begin{itemize}[leftmargin=*]
  \item \textbf{\textit{Automatic Speech Recognition.}}
Consistent with neural machine translation, automatic speech recognition (ASR) has benefited dramatically from non-autoregressive models. 
NAR ASR models can significantly speed up the decoding process but also suffer from lower recognition accuracy 
due to the failure of capturing target side dependency. 
The difference reflects in the processing unit, which is a unique characteristic in NAR ASR~\cite{higuchi2021comparative}. 
The models with token-level processing units need length prediction, while models with frame-level need not.
Thus, many NAR methods in neural machine translation cannot be directly used for ASR, but require specific modifications and designs, e.g., Iteration-based~\cite{chi2021align,chen2021align}, Audio-CMLM~\cite{chen2020non}, Imputer~\cite{chan2020imputer}, Mask-CTC~\cite{higuchi2020mask}, and Insertion-based~\cite{stern2019insertion,chan2019kermit} methods.
Besides, knowledge distillation is also an
effective skill for NAR ASR~\cite{gong2022knowledge,yoon2023inter}.
Considering that the most widely used CTC method in NAR ASR is under the assumption that there exists strong conditional independence between different token frame predictions, 
researchers have made considerable efforts to optimize the vanilla CTC-based model
~\cite{lee2021intermediate,nozaki2021relaxing,geng2021learning,deng2022improving,fan2021cass,higuchi2021improved,song2021non,yang2022improving,lu2023context,dingliwal2023personalization}.
Simultaneously, similar to the NAT method, the NAR ASR model can also benefit from pre-trained models, e.g., BERT~\cite{bai2021fast,yu2021non,higuchi2022bectra}. 
Besides, Higuchi~\etal~\cite{higuchi2021comparative} carry out a comparative study on NAR ASR to better understand this task. 
Recently, error correction for NAR ASR has also been explored~\cite{fan2022acoustic,futami2022non}.

\item \textbf{\textit{Summarization.}}
The summarization task is less subject to target side dependency modeling than neural machine translation since all the target output information is explicitly or implicitly included in the long text input.
As a result, NAR methods for the summarization task mainly alleviate the challenge of length prediction.
For instance, a Non-Autoregressive Unsupervised Summarization (NAUS) model has been proposed recently~\cite{liu2022learning}, which first performs an edit-based search towards a heuristically defined score and then generates a summary as a pseudo-ground-truth.
The authors also propose a length-control decoding approach for better target length prediction.
Furthermore, Liu~\etal~\cite{liucharacter} propose a Non-Autoregressive summarization model with Character-level length Control (NACC), which extends the length control algorithm to character level and achieves significant performance improvements on several datasets.

\item \textbf{\textit{Grammatical Error Correction.}}
Grammatical Error Correction (GEC) is an important NLP task that can automatically detect and correct grammatical errors within a sentence.
As most contents of a sentence are correct and unnecessary to be modified for the GEC task, the problem of lacking target side information can be effectively alleviated.
Thus, NAR methods are more feasible for this task.
Li~\etal~\cite{li2021tail} focus on the variable-length correction scenario for Chinese GEC.
They employ BERT to initialize the encoder and add a CRF layer on the initialized encoder, augmented by a focal loss penalty strategy to capture the target side dependency.
Besides, Straka~\etal~\cite{straka2021character} propose a character-based non-autoregressive GEC approach for Czech, German and Russian languages, which focuses on sub-word errors.
Shen~\etal~\cite{shen2022mask} propose a simple yet effective masking strategy to encourage the model to focus on the correct tokens and thus to better understand the sentence.

\item \textbf{\textit{Dialogue.}}
Dialogue generation has achieved remarkable progress in the last few years, and many methods have been proposed to alleviate the notorious problem of diversity~\cite{kulikov2019importance}. 
However, due to their autoregressive generation strategy, these dialogue generation models suffer from low inference efficiency for generating informative responses.
Inspired by the advances made in NAT~\cite{bao2019non,sun2019fast}, NAR models are adopted in dialogue generation to lower the inference latency, where the response length is predicted in advance.
Han~\etal~\cite{han2020non} apply the NAR model to model the bidirectional conditional dependency between contexts (x) and responses (y).
They also point out that NAR models can produce more diverse responses. 
Zou~\etal~\cite{zou2021thinking} propose a concept-guided non-autoregressive method for open-domain response generation, which customizes the Insertion Transformer to complete response and then facilitates a controllable and coherent dialogue.
These NAR models for dialogue generation can significantly improve response generation speed.
Besides, NAR methods can improve task-oriented dialogue systems by enhancing the spoken language understanding sub-task~\cite{cheng2021effective,cheng2022capture}.

\item \textbf{\textit{Text Style Transfer.}}
Autoregressive models have been widely used in unsupervised text style transfer. 
Despite their success, they suffer from high inference latency and low content preservation problems. 
Several works explore non-autoregressive (NAR) decoding to alleviate these problems.
Ma~\etal~\cite{ma2021exploring} first directly adapt the common training scheme from the AR counterpart in their NAR method and then propose to enhance the NAR decoding from three perspectives: knowledge distillation, contrastive learning, and iterative decoding. 
They also explore the potential reasons why these methods can narrow the performance gap with AR models.
Huang~\etal~\cite{huang2021nast} point out that the autoregressive manner might generate some irrelevant words with strong styles and ignore part of the source sentence content.
They propose a NAR generator for unsupervised text style transfer (NAST), which effectively avoids irrelevant words by alignment information. 
NAST can dramatically improve transfer performance with efficient decoding speed.

\item \textbf{\textit{Controllable Text Generation.}}
Controllable Text Generation (CTG) is an emerging area in the field of natural language generation~\cite{zhang2022survey}. It aims to generate texts that meet certain controllable constraints as humans wish reliably. These constraints are generally task-specific, and CTG can be exploited in various tasks. A few recent works have begun to explore CTG on NAR models.
Agrawal~\etal~\cite{agrawal2021non} introduce a non-autoregressive approach for controllable text simplification, where the model iteratively edits an input sequence and incorporates lexical complexity information into the refinement process to generate simplifications.
Iso~\etal~\cite{iso2022autotemplate} propose AutoTemplate for lexically constrained text generation task, which decomposes the generation process into two steps, template generation and lexicalization, by converting the input and output formats.
Li~\etal~\cite{lidiffusion} apply the diffusion model to six fine-grained controllable tasks. Surprisingly, their method doubles the control success rate of prior methods and is competitive with strong baseline methods that require additional training (fine-tuning).
Recently, Kumar~\etal~\cite{kumar2022gradient} propose MUCOLA, a sampling strategy that flexibly combines pre-trained language models with differentiable constraints.
Evaluation with several CTG tasks proves that MUCOLA can achieve excellent performance on toxicity avoidance, sentiment control, and keyword-guided generation.

\item \textbf{\textit{Image Captioning.}}
Image captioning is the task of generating natural language captions for given images.
In recent years, many researchers have brought NAR models into image captioning, and their introduced methods have achieved very appealing progress \cite{fei2019fast,gao2019masked,guo2021non}. Motivated by Levenshtein Transformer~\cite{gu2019levenshtein}, Wang~\etal~\cite{wang2022explicit} propose TIger for image captioning, which consists of three modules, i.e., Inserter, $\text{Tagger}_{\text{del}}$ and $\text{Tagger}_{\text{add}}$ to achieve explicit caption editing. 
Besides, Fei~\etal~\cite{fei2022efficient} customize the shared encoder~\cite{hao2021multi,zhou2022confidence} and a NAR decoder for image captioning to improve the modeling capacity of the AR model.
Fei~\etal~\cite{fei2022uncertainty} also introduce a novel uncertainty-aware framework that leverages an Insertion Transformer-based~\cite{stern2019insertion} structure to generate image captions from easy to difficult non-autoregressively and an uncertainty-adaptive beam search technique to speed up the decoding further.
Chen~\etal~\cite{chenlearning} additionally explore the potential of a fully NAR model for image captioning, in which a Discrete Mode Learning (DML) paradigm is employed to alleviate the mode collapse problem.

\item \textbf{\textit{Question Answering.}}
NAR components also serve a vital role in question answering. 
To solve the exposure bias problem when using AR models in hybrid tabular-textual question answering, Zhang~\etal~\cite{zhang2022napg} propose a non-autoregressive program generation framework, which can generate complete program tuples in parallel and help address the error accumulation issue, and thus can boost both the performance and efficiency. 
Wang~\etal~\cite{wang2022kecp} propose Knowledge Enhanced Contrastive Prompt-tuning (KECP) for Extractive Question Answering (EQA).
They transform the task into a NAR Masked Language Modeling (MLM) generation problem without additional pre-training stages. 
Experiments on multiple benchmarks demonstrate the effectiveness of the proposed strategy.

\end{itemize}

\subsection{Semantic Parsing}
Compared with the non-autoregressive text generation tasks, non-autoregressive semantic parsing relies more on the length prediction mechanism, in which minor differences can lead to entirely different results. 
Several NAR models applied to semantic parsing are inspired by CMLM~\cite{ghazvininejad2019mask} but with better length prediction mechanisms. 
Babu~\etal~\cite{babu2021non} study the potential limitations of the original CMLM when applied for semantic parsing and designed a new LightConv Pointer model to improve it,
where the target length is computed by a separate module of multiple layers of CNNs with gated linear units.
They also use label smoothing to avoid the easy over-fitting in length prediction. 
During inference, iterative refinement does not bring many benefits to task-oriented semantic parsing, and thus only one step is applied.
Shrivastava~\etal~\cite{shrivastava2021span} design Span Pointer Networks based on CMLM with a span prediction mechanism to decide the target length.
The length module of semantic parsing merely needs frame syntax to perform span prediction, while text generation requires both syntax and semantics.

\subsection{Text to Speech}
Significant progress has also been made in the non-autoregressive text to speech (NAR TTS) task. 
Ren~\etal~\cite{ren2019fastspeech} point out three main problems in autoregressive TTS compared with the non-autoregressive fashion, i.e., the speed of the inference stage is slow, the generated speech is not robust, and the generated speech is unable to be controlled.
Accordingly, they present a model based on Transformer in a non-autoregressive manner to alleviate the above three problems. 
Besides, one-to-many (O2M) mapping problem is typical in NAR TTS since differences lie in human speaking greatly. Many other NAR TTS models are also proposed to alleviate this problem and improve speech quality.
Peng~\etal~\cite{peng2020non} propose ParaNet, which extracts attention from the autoregressive TTS model and then re-defines the alignment.
Lu~\etal~\cite{lu2021vaenar} apply the variational auto-encoder structure to model the alignment information with a latent variable and further use the attention-based soft alignment strategy.
Shah~\etal~\cite{shah2021non} propose a NAR model by replacing the attention module of the conventional attention-based TTS model with an external duration model for low-resource and highly expressive speech.
Besides, a very deep VAE model with residual attention also benefits the NAR TTS~\cite{liu2021vara}. 
Notice that the above models may need a teacher model to guide their learning. 
Lee~\etal~\cite{lee2020bidirectional} propose a bidirectional inference variational auto-encoder to rely less on the teacher model and meanwhile without decreasing the performance.
Since over-smoothing is a severe problem that harms the performance of NAR TTS models, many works focus on alleviating this problem. Ren~\etal~\cite{ren2022revisiting} summarize these methods into the two categories, i.e., simplify data distributions~\cite{wang2017tacotron,lancucki2021fastpitch}, which provides more conditional input information,  and enhance modeling methods~\cite{guo2022multi,kim2020glow}, which try to enhance the model capacity to fit the complex data distributions. Ren~\etal~\cite{ren2022revisiting} combine these two methods to improve the performance of NAR TTS further.
The diversity problem of TTS is also explored in recent work.
Bae~\etal~\cite{Bae2022hierarchical} propose a variational autoencoder with the hierarchical and multi-scale structure for NAR TTS (HiMuV-TTS) to improve the diversity of generated speech.
As most parallel end-to-end TTS models fail to disentangle general prosody features from the speech, 
Li~\etal~\cite{li2022cross} introduce a cross-utterance conditional VAE (CUC-VAE) system to achieve better naturalness and more prosody diversity.
Besides, Liu~\etal~\cite{liu2022controllable} propose Controllable and LOssless Non-autoregressive End-to-end TTS (CLONE) to model the general prosody effectively.

\subsection{Speech Translation}
Much progress has also been made in speech translation along with the development of NAR ASR models mentioned in section \ref{sec:Task-Specific}.
Many NAR ASR models are applicable for end-to-end speech translation~\cite{post2013improved} by completing the automatic speech recognition and machine translation stages simultaneously. 
Since speech translation resembles text translation, effective strategies applied in text translation are also introduced to speech translation. 
In seeing the success of connectionist temporal classification (CTC) on machine translation~\cite{shu2020latent}, Chuang~\etal~\cite{chuang2021investigating} propose CTC-based speech-to-text translation model. 
They construct an auxiliary speech recognition task based on CTC to further improve performance.
Inaguma~\etal~\cite{inaguma2021orthros} propose Orthros to jointly train the NAR and AR decoders on a shared speech encoder, which is similar to sharing encoder structure in machine translation~\cite{hao2021multi}. 
Besides, a rescoring mechanism is proposed for Orthros~\cite{inaguma2021non}, in which an auxiliary shallow AR decoder is introduced to choose the best candidate. 
On the NAR side, they use CMLM and a CTC-based model as NAR decoders, denoted as Orthros-CMLM and Orthros-CTC, respectively.
Such muti-decoder is also widely used for speech translation~\cite{dalmia2021searchable,shi2021highland}, which is a two-pass decoding method that decomposes the overall task into two sub-tasks, i.e., ASR and machine translation. 
Inaguma~\etal~\cite{inaguma2021fast} propose Fast-MD, where the hidden intermediates are generated in a non-autoregressive manner by a Mask-CTC model. They also introduce a sampling prediction strategy to reduce the mismatched training and testing.

\subsection{Diffusion Models}
The diffusion model is first proposed in \cite{sohl2015deep}, which estimates the data distribution $X_0 \in \mathbb{R}^{d}$ through a series of latent variables $X_T \cdots X_0$ as Markov chain,
with each variable $X_i \in \mathbb{R}^{d}$ and $X_T$ a Gaussian noise. During training, the diffusion model defines a forward process that constructs the intermediate latent variables $X_1 \cdots X_T$ by incrementally adding Gaussian noise to data $X_0$ until it turns to approximate a Gaussian at diffusion step $T$.
\begin{figure}[tb] 
\centering
\includegraphics[scale=0.32]{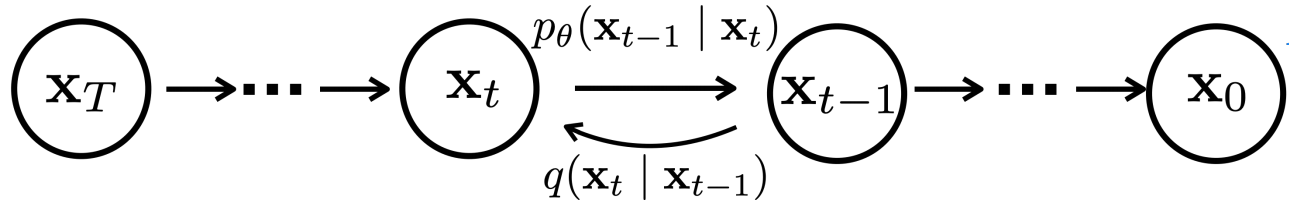}
\vspace{-0.2cm}
\caption{The forward and reverse process of a diffusion model~\cite{ho2020denoising}.}
\label{fig:diffusion}
\vspace{-0.5cm}
\end{figure} 
As shown in Figure~\ref{fig:diffusion}, $q(X_{t}|X_{t-1})$ denotes the forward process and $p_\theta(X_{t-1}|X_t)$ refers to the reverse process estimated by $p_\theta$.
During inference, diffusion models adopt a non-autoregressive manner to denoise from $X_T$ (a Gaussian) to $X_0$ (target data).
Early research mainly explores their effectiveness on continuous data, such as images and audio generation~\cite{ho2020denoising,kong2020diffwave,mittal2021symbolic,nichol2021improved,austin2021structured,sahariaphotorealistic,ramesh2022hierarchical,nichol2022glide}. 
Very few works investigate diffusion models with discrete state spaces, such as text generation~\cite{nichol2022glide,hoogeboom2021argmax,hoogeboomautoregressive}.
Among which, Savinov~\etal~\cite{savinovstep} learn from the vanilla diffusion model and propose SUNDAE, a step-unrolled denoising autoencoder that introduces a specific corruption function during training. 
Unlike previous diffusion models, SUNDAE converges in fewer iterations during inference. More recently, Li~\etal~\cite{lidiffusion} propose Diffusion-LM, which adopts continuous latent representations and efficient gradient-based methods for controllable text generation. 
Gong~\etal~\cite{gong2022diffuseq} further extend the diffusion model to more text generation tasks, which achieves promising performance. 
Yu~\etal~\cite{yu2022latent} combine Energy-Based Models (EBMS) and the diffusion model for interpretable text modeling. 
Reid~\etal~\cite{reid2022diffuser} propose DIFFUSER, a new edit-based generative model for text generation based on denoising
diffusion models. 
They also confirm the effectiveness of the diffusion model on machine translation, summarization, and style transfer.
There are also existing works that draw the connection between non-autoregressive methods and the diffusion model. 
In fact, Bert~\cite{kenton2019bert} and generative masked language models~\cite{ghazvininejad2019mask,wang2019bert} can also be viewed as diffusion models. They just adopt the different training methods to model the denoising process~\cite{austin2021structured}. Gong~\etal~\cite{gong2022diffuseq} conduct a theoretical analysis to reveal the connection between their diffusion model DIFFUSEQ and
non-autoregressive models. They claim that DIFFUSEQ can be seen as a more generalized form of an iterative non-autoregressive model. As mentioned in Section~\ref{sec:iteration}, each denoising process from $X_T$ to $X_0$ of the diffusion model serves as a decoding iteration.

\subsection{Others}
In addition to the success of NAR methods on the above-mentioned tasks, many researchers have conducted a pilot study on other scenarios. 
Information Extraction (IE) also benefits from the non-autoregressive technique. In essence, the facts in plain text are unordered, but the AR models need to predict the following fact conditioned on the previously decoded ones.
Yu~\etal~\cite{yu2021maximal} propose a novel non-autoregressive framework, named  MacroIE, for OpenIE, which treats IE as a maximal clique discovery problem and predicts the fact set at once to relieve the burden of predicting fact order.
For Video Generation (VG), Yu~\etal~\cite{yu2021generating} propose a dynamics-aware implicit generative adversarial network (DIGAN) for non-autoregressive video generation, which greatly increases inference speed via parallel computing of multiple frames. 
For Voice Conversion (VC), Hayashi~\etal~\cite{hayashi2021non} extend the FastSpeech2 model in NAR TTS to the voice conversion task and introduce a convolution-augmented Transformer (Conformer).
The proposed method can learn both local and global context information of the input sequence and extend variance predictors to variance converters to transpose the prosody components of the source speaker.
A fully non-autoregressive many-to-many voice conversion method is also presented in~\cite{chen2022streaming}, which includes a streaming transformer-based acoustic model and a streaming vocoder. Wang~\etal~\cite{wang2022fastlts} propose FastLTS for unconstrained lip-to-speech synthesis. They use a GAN-based vocoder along with adversarial training to improve audio quality and adopt a fully parallelized architecture with a non-autoregressive decoder and vocoder to improve inference efficiency.
Besides, self-supervised speech representations are effective in various speech applications. 
However, existing representation learning methods generally rely on the autoregressive model, leading to low inference efficiency. Liu~\etal~\cite{liu2020non}
propose Non-Autoregressive Predictive Coding (NPC) to learn speech representations in a non-autoregressive manner by only considering local dependencies of speech, which can significantly improve inference speed.
The decoding efficiency of full-line code completion can also benefit from NAR models~\cite{liu2022non}, where a syntax-aware sampling strategy is leveraged to improve the performance. The authors further point out that the dependency on target tokens in code completion is weaker, which is profit for NAR modeling.
Barezi~\etal~\cite{barezi2020study} make an attempt to adopt the NAR model in multi-label learning for extreme classification tasks. Their designed non-autoregressive latent variable model significantly outperforms the autoregressive baselines.
Feng~\etal~\cite{feng2022multi} propose Multi-scale Attention Normalizing Flow(MANF), a novel non-autoregressive deep learning model for time series forecasting tasks. MANF can avoid the influence of cumulative error and meanwhile reduce the time complexity.
\section{Conclusion and Outlooks}
\label{sec:conclusion}
This paper reviews the development of non-autoregressive methods in neural machine translation and other related tasks. 
We first summarize the main challenge encountered in NAT research.
Then, we structure existing solutions from different perspectives, including data manipulation, modeling, criterion, decoding, and benefiting from pre-trained models, along with a discussion on their effectiveness and inference speed.
Besides, we present an overview of the applications of NAR methods in extensive tasks, e.g., summarization, semantic parsing, text to speech, and speech translation.
We hope this survey can help researchers and engineers better understand the non-autoregressive techniques and choose suitable strategies for their application tasks.

Although impressive progress has been made on non-autoregressive models, there still exist some open problems: 
\begin{itemize}
    \item KD is the most effective method utilized in NAR models, which depends on pre-training an AR model in advance. However, how to release this condition and improve the performance of NAR models on raw datasets are worthy of further consideration.
    \item Although iteration-based models have been proposed to help capture the target side contextual dependency in multiple decoding steps and achieved comparable performance with AR models, their speedup w.r.t AR models will diminish when decoding with large batch sizes~\cite{kasai2020deep,helcl2022non}. Therefore, more attention should be paid to mitigate the challenge mentioned above under the framework of fully NAT models.  
    \item Reasonable training objectives are critical for capturing the target side dependencies for NAR models. Recently, Huang~\etal~\cite{huang2022learning} point out that simply training NAT models by maximizing the likelihood can lead to an approximation of marginal distributions but drops all dependencies between tokens, and they revisit the previous success~(including some advanced criterion introduced in Section~\ref{sec:criterion}) in a unified framework. Thus, how to design suitable training objectives is worth further exploration.
    \item AR models are generally applied to various application scenarios, including bilingual and multilingual, high-resource and low-resource, etc. However, most applications of NAR models are limited to the bilingual scenario until now. Therefore, to expand the impact of NAR models, it is worthy of applying NAR to more application scenarios.  
    \item In recent years, considerable efforts have been made to enhance autoregressive models with powerful pre-training techniques and models, with impressive performance being achieved. However, only very few papers apply these powerful pre-trained models to help NAR models~\cite{guo2020incorporating,jiang2021improving}, and there is only a preliminary exploration of the pre-training techniques for NAR models~\cite{qi2021bang,li2022universal,li2022elmer}. Thus, it is promising to explore pre-training methods for non-autoregressive generation and other related tasks.  
\end{itemize}

\bibliographystyle{IEEEtran}
\bibliography{IEEEtran}

\begin{IEEEbiography}[{\includegraphics[width=1in,height=1.25in,clip,keepaspectratio]{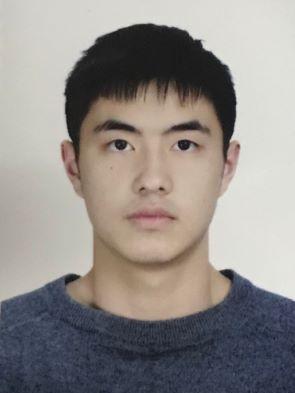}}]{Yisheng Xiao} is now a graduate student at the natural language processing laboratory, Soochow University, supervised by Prof. Min Zhang. He received his Bachelor's degree, majoring in software engineering, from the same place in 2021. His research interests lie in Natural Language Processing, especially in neural machine translation, non-autoregressive methods, efficient text generation, and pre-trained language models.

\end{IEEEbiography}

\vskip -3\baselineskip plus -1fil

\begin{IEEEbiography}[{\includegraphics[width=1in,height=1.25in,clip,keepaspectratio]{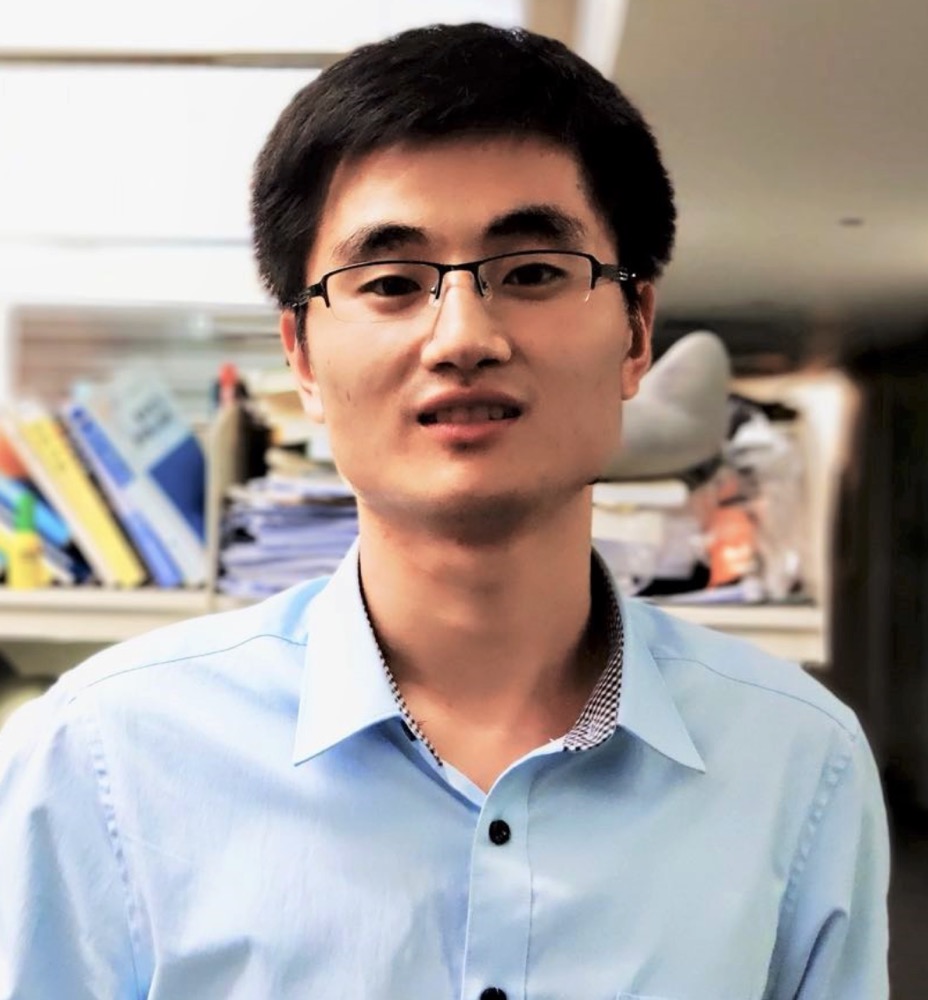}}]{Lijun Wu} is a Senior Researcher of Machine Learning Group in Microsoft Research Asia (MSRA). 
He got his Ph.D. degree from Sun Yat-sen University (SYSU) in 2020, a member of the joint Ph.D. program between SYSU and MSRA.
He received MSRA Ph.D. Fellowship in 2018. 
His researches focus on Deep Learning, Natural Language Processing, Multimodality Learning, and Medical Health. 
\end{IEEEbiography}

\vskip -3\baselineskip plus -1fil

\begin{IEEEbiography}[{\includegraphics[width=1in,height=1.25in,clip,keepaspectratio]{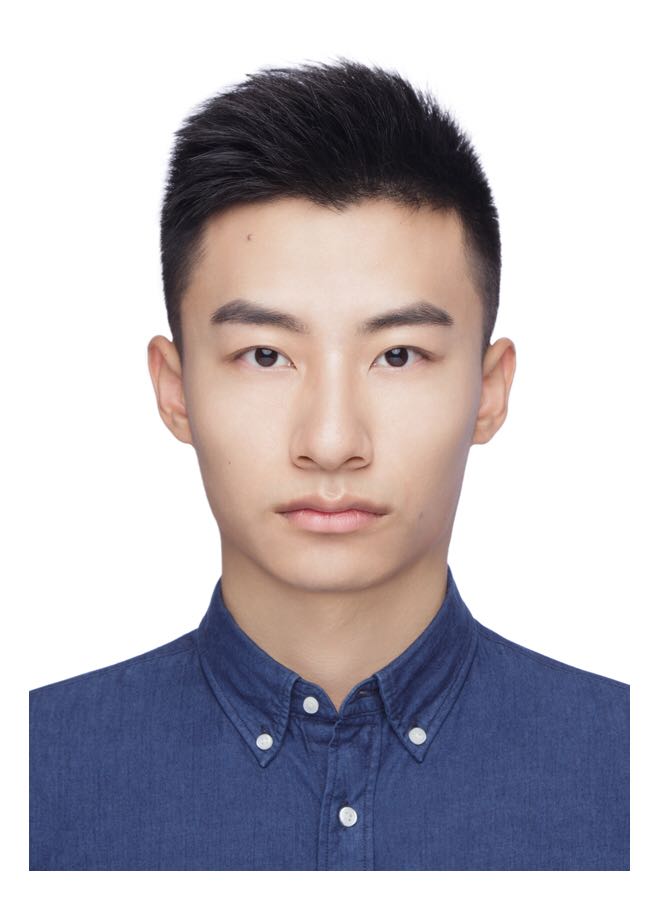}}]{Junliang Guo} is a Researcher of the Machine Learning group in Microsoft Research Asia (MSRA). He received his Ph.D. degree in computer science from the University of Science and Technology of China (USTC) in 2021. His research interests lie in the general area of machine learning, specifically in sequence modeling and representation learning, as well as their applications in natural language processing and heterogeneous types of data such as networks. 
\end{IEEEbiography}

\vskip -3\baselineskip plus -1fil

\begin{IEEEbiography}[{\includegraphics[width=1in,height=1.25in,clip,keepaspectratio]{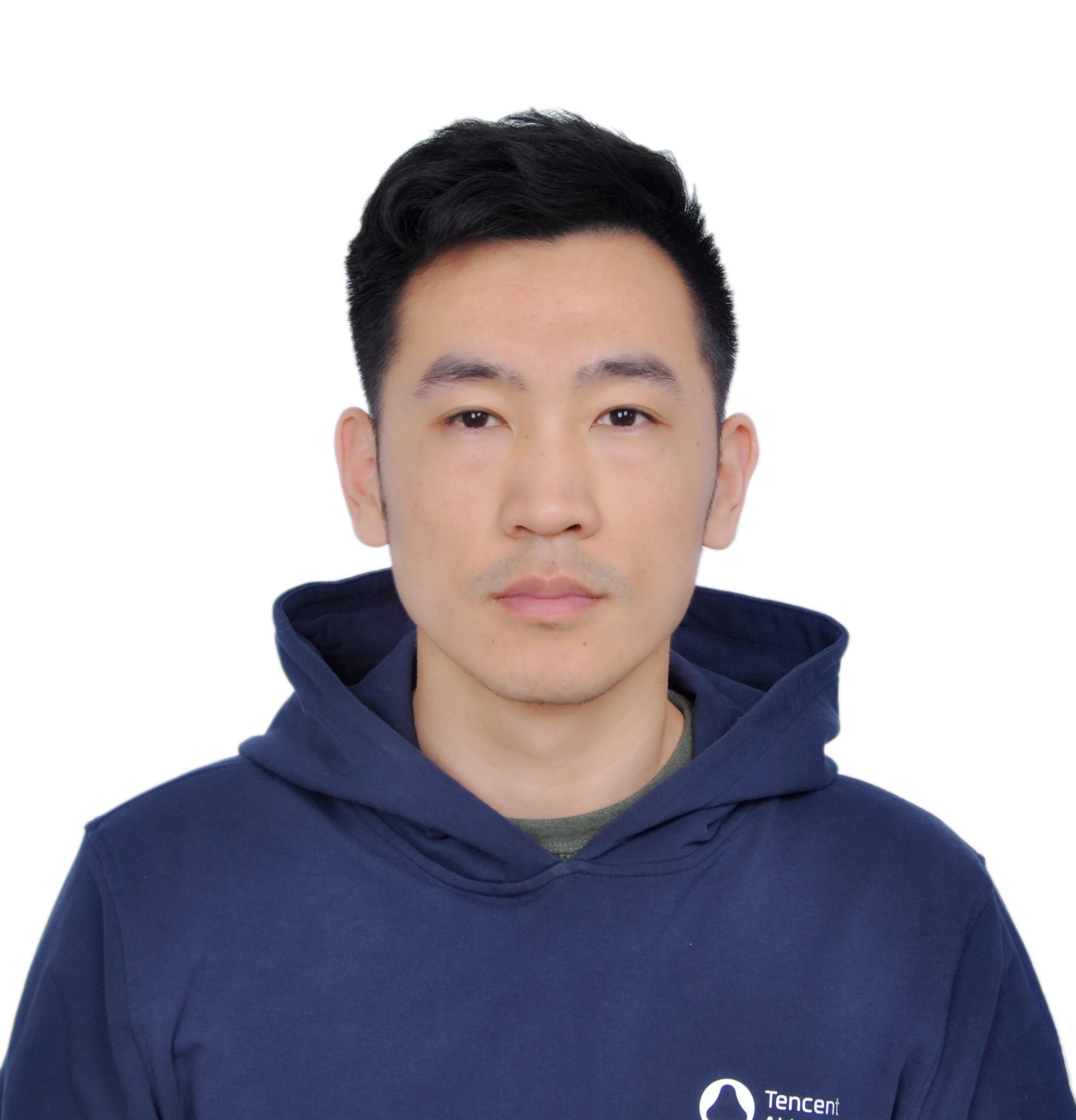}}]{Juntao Li} is now an associate professor at the Institute of Artificial Intelligence, Soochow University. 
Before that, he obtained a doctoral degree from Peking University in 2020.
He is now working on pre-trained language models, text generation, and dialogue systems.
\end{IEEEbiography}

\vskip -3\baselineskip plus -1fil

\begin{IEEEbiography}[{\includegraphics[width=1in,height=1.25in,clip,keepaspectratio]{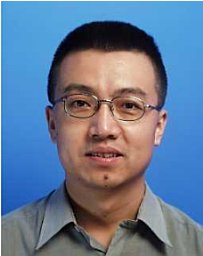}}]{Min Zhang} is a Distinguished Professor at Soochow University (China). 
He received his bachelor’s degree and Ph.D. degree from the Harbin Institute of Technology in 1991 and 1997, respectively. 
His current research interests include machine translation, natural language processing, and machine learning.
He has authored 150 papers in leading journals and conferences and has co-edited 10 books that were published by Springer and IEEE. 
\end{IEEEbiography}

\vskip -3\baselineskip plus -1fil

\begin{IEEEbiography}[{\includegraphics[width=1in,height=1.25in,clip,keepaspectratio]{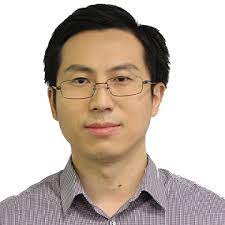}}]{Tao Qin} received the Ph.D. degree and Bachelor degree both from Tsinghua University. 
He is a Senior Member of ACM and IEEE and a Senior Principal Research Manager in Machine Learning Group, Microsoft Research Asia. 
His research interests include machine learning (with the focus on deep learning and reinforcement learning), artificial intelligence (with applications to language understanding and computer vision).
\end{IEEEbiography}

\vskip -3\baselineskip plus -1fil

\begin{IEEEbiography}[{\includegraphics[width=1in,height=1.25in,clip,keepaspectratio]{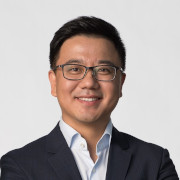}}]{Tie-Yan Liu} received the Ph.D. degree and Bachelor degree both from Tsinghua University. 
He is an Assistant managing director of Microsoft Research Asia, leading the machine learning research area.
He is an adjunct/honorary professor at Carnegie Mellon University (CMU).
He has published 200+ papers in refereed conferences and journals, e.g., ICML, KDD, NeurIPS, with 30000+ citations. He is a fellow of IEEE and ACM.
\end{IEEEbiography}

\clearpage
\appendix
\section*{Overview of Improving Methods for NAT}
\begin{figure*}[t] 
\centering
\includegraphics[scale=0.70]{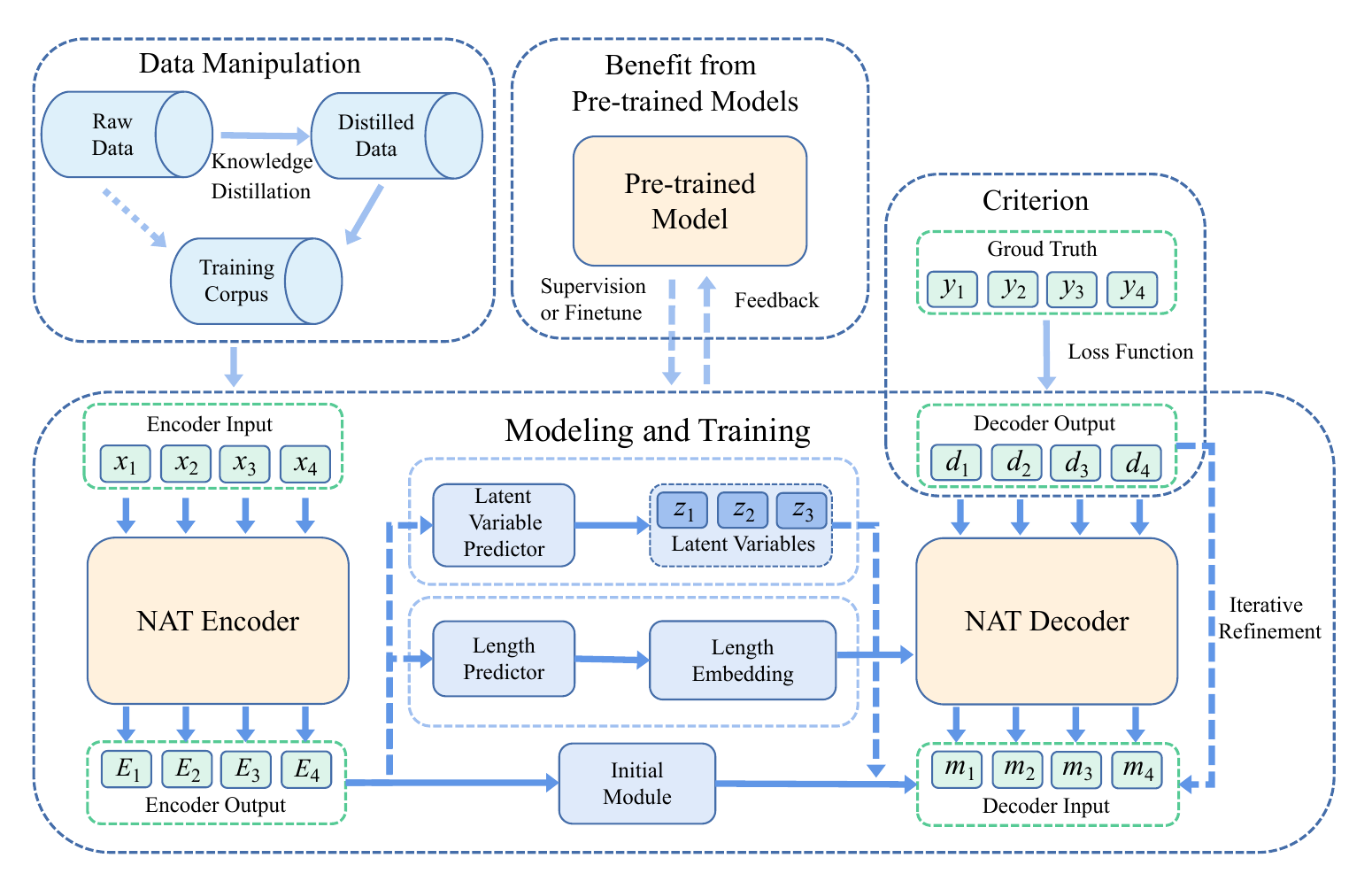}
\caption{The overall framework of various components for improving NAT models. The dashed arrow denotes this part is not applied in all NAT models.
Different knowledge distillation methods will be applied in Data Preparing part. Length Predictor is involved in most NAT models, and Latent Variable Predictor is applied in latent variable-based models. The initial Module is used to initialize the Decoder Input, such as soft copy, source copy, partial target tokens, etc.}
\label{fig:MODEL} 
\end{figure*}

To have a clear overview of improving methods for NAT, we show the general framework and the data flow of various NAT models in Figure~\ref{fig:MODEL}, which contain different components such as the data preparation, NAT encoder, and NAT decoder. 

\section*{Representative Modeling Methods}
\begin{figure*}[!ht] 
\centering
\subfigure[An illustration of two iteration-based methods introduced in Section~\ref{sec:iteration}.]{
\includegraphics[scale=0.63]{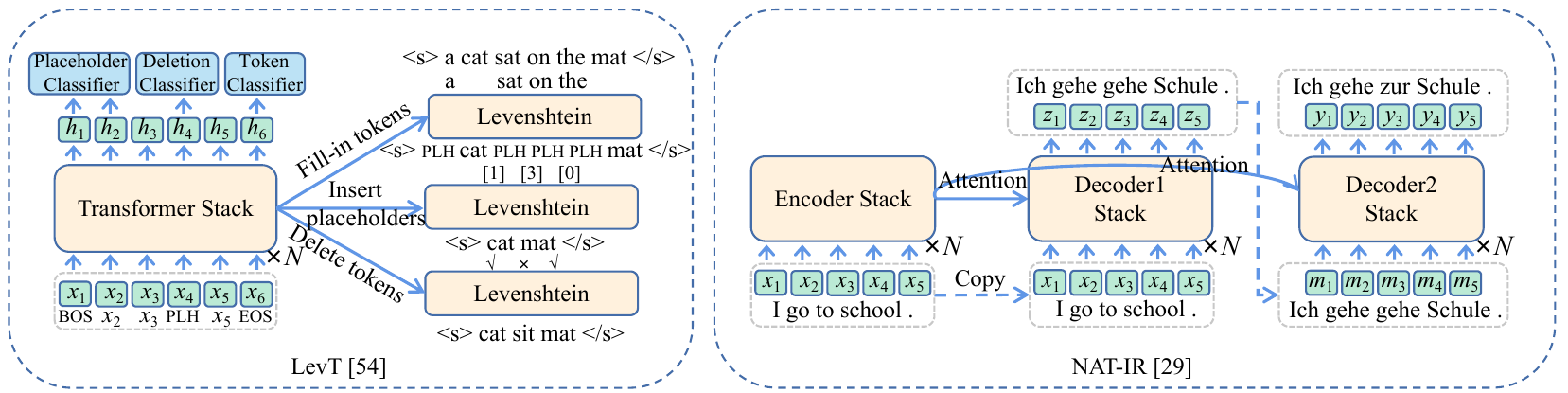}
}
\subfigure[An illustration of two latent variable-based methods introduced in Section~\ref{sec:latent}.]{
\includegraphics[scale=0.63]{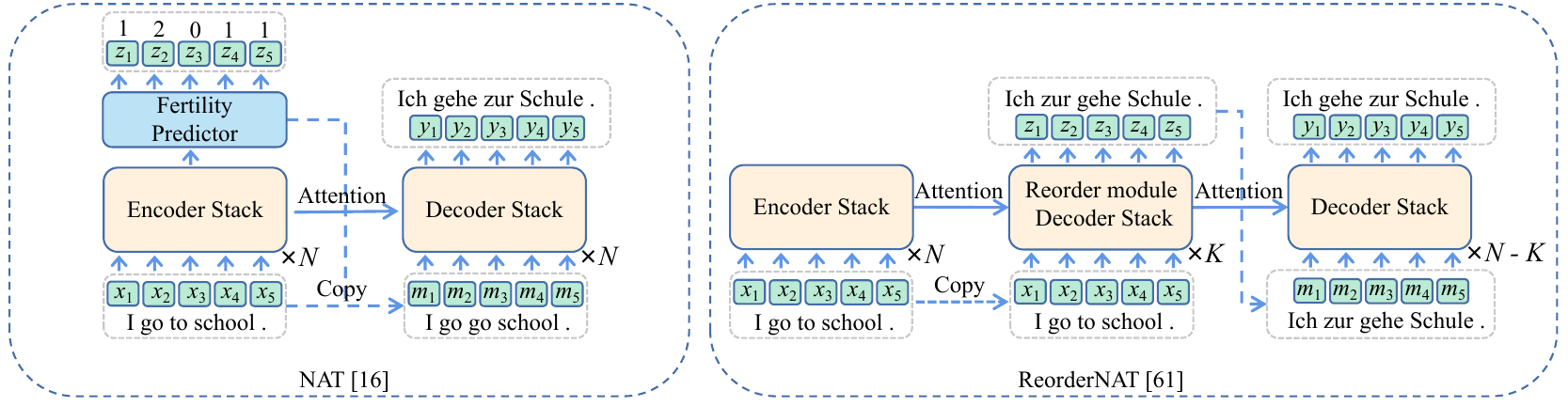}
}
\subfigure[An illustration of two enhancements-based methods introduced in Section~\ref{sec:enhancement}.]{
\includegraphics[scale=0.63]{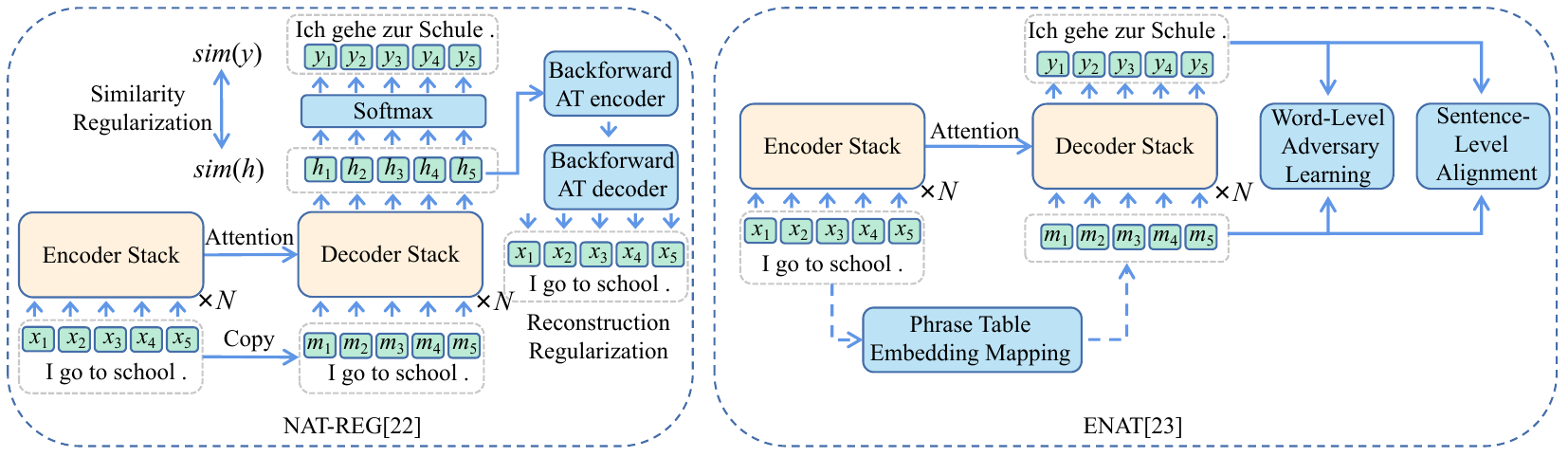}
}
\caption{We show the model structures of two iteration-based methods, e.g., two iteration-based methods, NAT-IR~\cite{lee2018deterministic} and LevT~\cite{gu2019levenshtein};  two latent variable-based methods, e.g., NAT~\cite{gu2018non} and ReorderNAT~\cite{ran2021guiding}, where the fertility predictor and reorder module are applied to predict the latent variables; and two enhancements-based model, NAT-REG~\cite{wang2019non} and ENAT~\cite{guo2019non}, and the corresponding enhancement module is also given. }
\label{fig:modeling} 
\end{figure*}

Figure~\ref{fig:modeling} presents a few representative modeling methods mentioned in Section~\ref{sec:model}. 

\section*{Criterion}
\begin{figure*}[t] 
\centering
\includegraphics[scale=0.65]{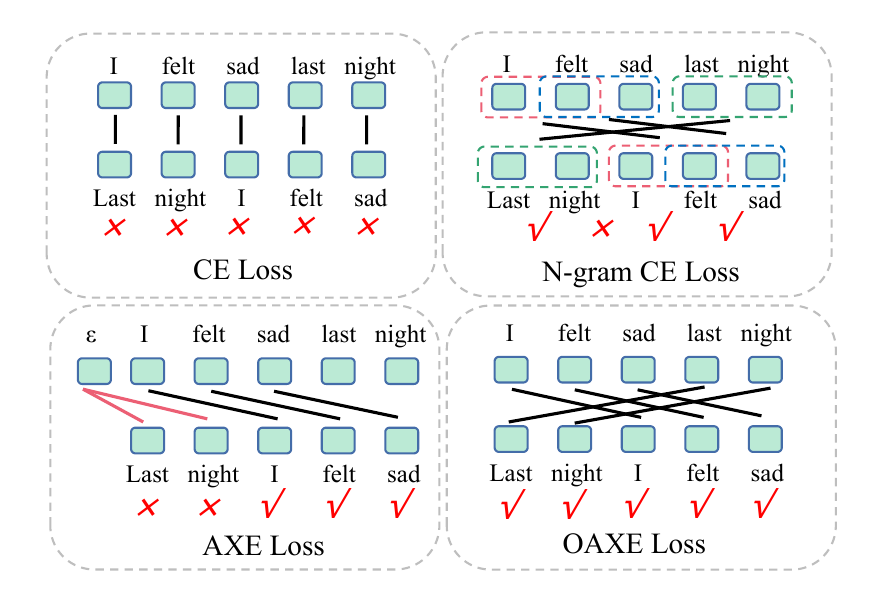}
\caption{An illustration of different loss functions. e.g., Model prediction: \textbf{Last night I feel sad} and Ground-truth: \textbf{I feel sad last night}. Traditional CE loss will give a penalty to all tokens. N-gram CE loss only finds a two-gram \textbf{night I} unreasonable.
AXE loss finds the best possible monotonic alignment and penalizes unaligned tokens, denoted as \textbf{$ \epsilon $}, while O\scriptsize{A}\footnotesize{XE} loss removes the order errors and give no penalty to this prediction.}
\label{fig:loss} 
\end{figure*} 
Figure~\ref{fig:loss} compares different loss functions in NAT models.

\section*{Illustration of Decoding Strategies}
\begin{figure*}[!th] 
\centering
\includegraphics[scale=0.65]{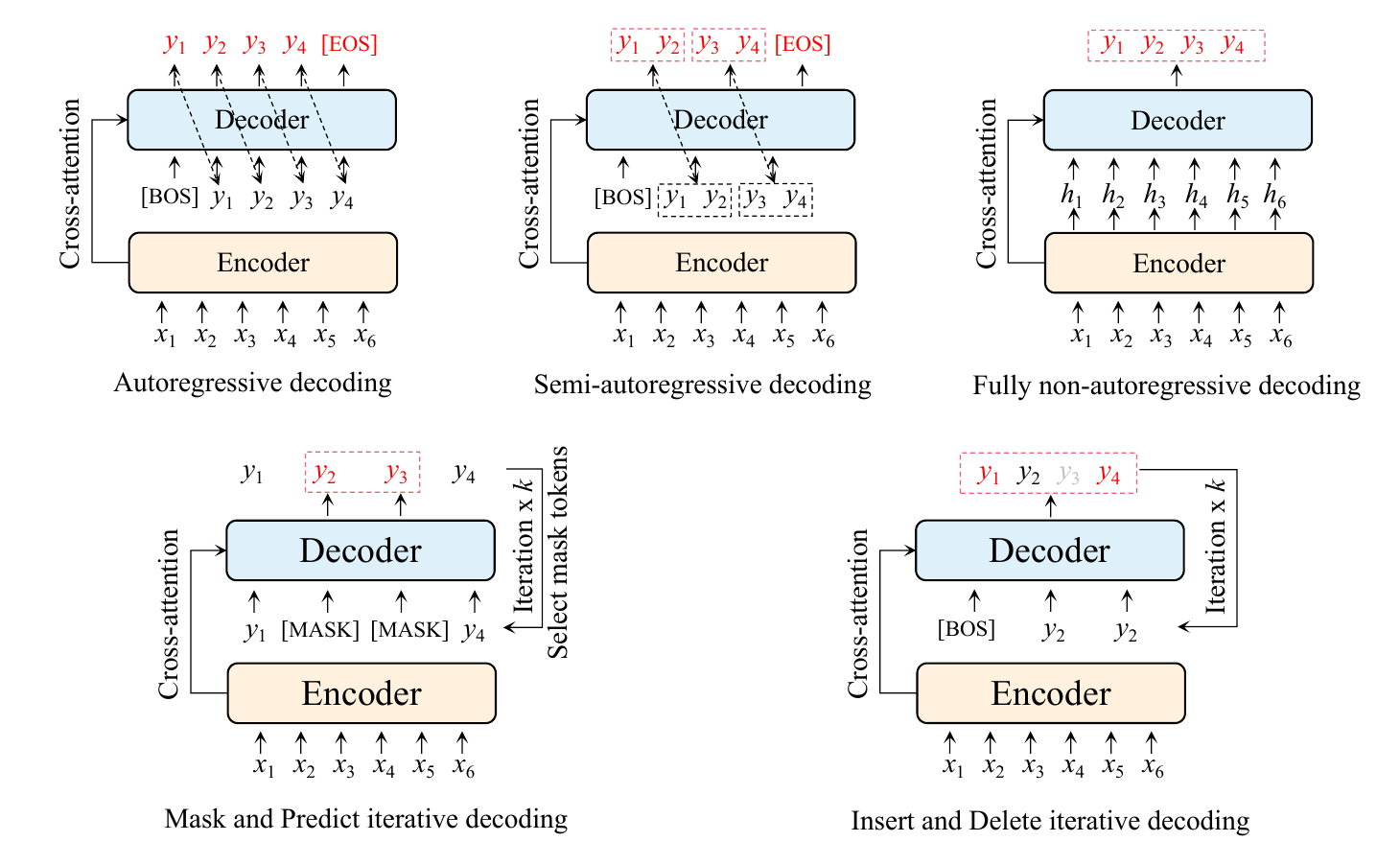}
\caption{An illustration of several different decoding strategies introduced in Section~\ref{sec:decoding}, e.g., autoregressive and semi-autoregressive decoding methods generate tokens in a left-to-right order, fully non-autoregressive and mask and predict iterative decoding methods predict all/part of target tokens in parallel (tokens marked by red), insert and delete iterative decoding method generate tokens with insertion (tokens marked by red) and deletion (tokens marked by grey) operations. }
\label{fig:decoding} 
\end{figure*}

\begin{figure*}[ht] 
\centering
\includegraphics[scale=0.8333]{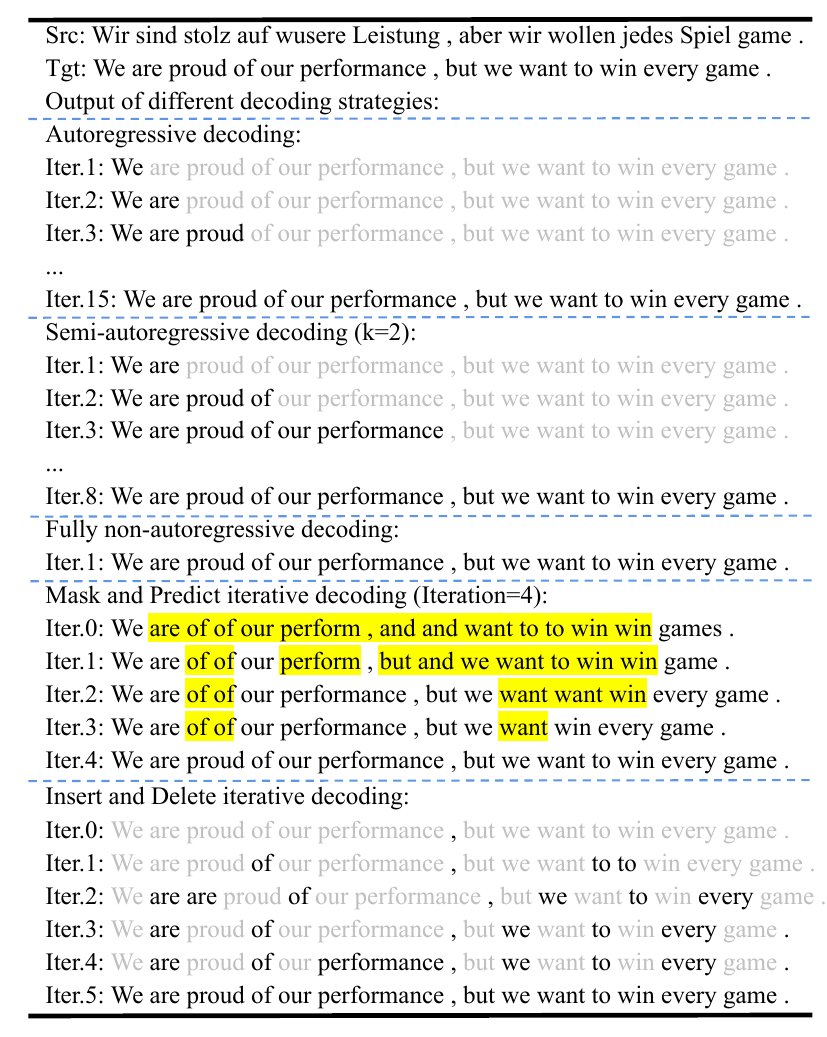}
\caption{Cases of several different decoding strategies discussed in this paper. Texts marked in yellow denote the content that will be masked and generated in next iteration.}
\label{fig:example} 
\end{figure*}

We show several typical decoding strategies in Figure~\ref{fig:decoding} and their detailed decoding process with a specific example in Figure \ref{fig:example}, including autoregressive decoding, semi-autoregressive decoding, fully non-autoregressive decoding, mask and predict iterative decoding, and insert and delete iterative decoding.

\section*{Comparisons between Existing Methods}
\label{app:comparisons}
\begin{figure*}[htp] 
\centering
\includegraphics[scale=0.55]{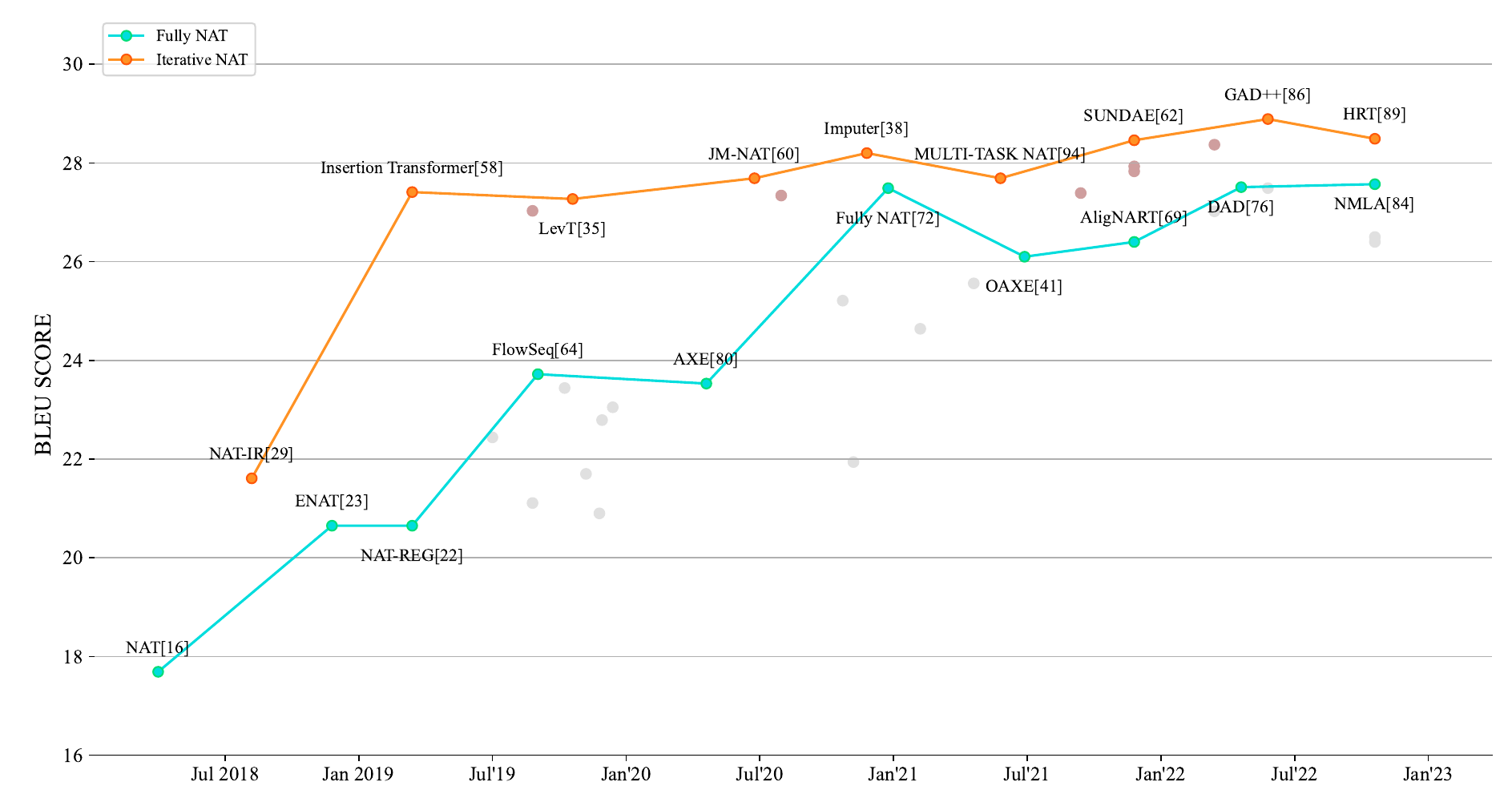}
\caption{BLEU v.s. Speedup. BLEU score is reported on the WMT14 EN$\to$DE test set. Speedup is reported for NAT models compared with corresponding AT models. Dotted lines denote the different scores of iterative models achieved with different iterations.
Note that the ideal model should appear in the top right-hand corner.}
\label{fig:bleuvsspeed} 
\end{figure*}

\begin{figure*}[ht] 
\centering
\includegraphics[scale=0.56]{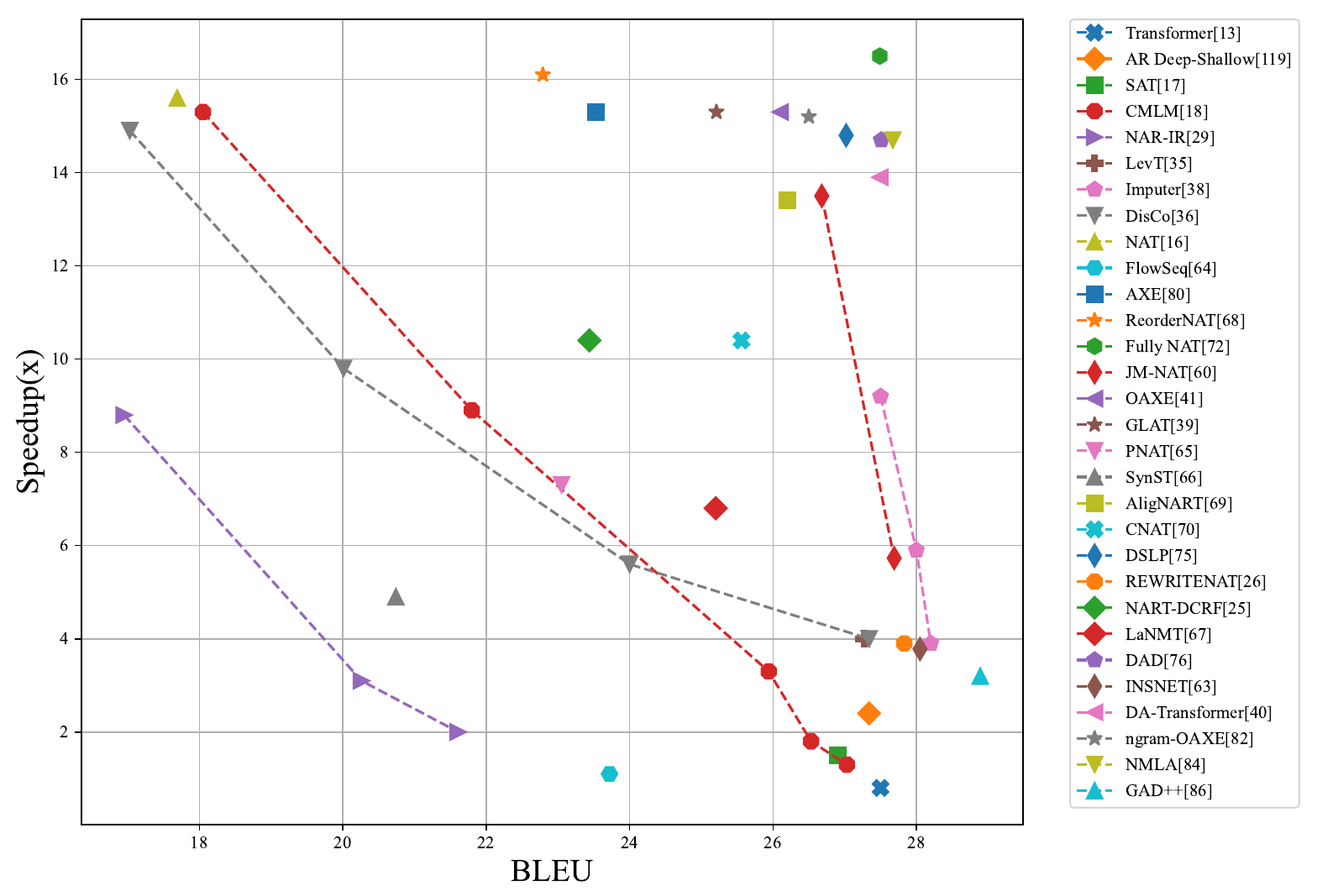}
\caption{The evolution of BLEU scores on WMT14 EN$\to$De translation by the time of Fully NAT and Iterative NAT. Note that the performance of iterative NAT models is commonly better than fully NAT models at different stages, but the gap is narrowing with the development of NAT models.}
\label{fig:frmework} 
\end{figure*}

In the paper, we mainly present the NAT works with their performances on the WMT14 English$\to$German translation task. Here we give a broader comparison of WMT14 English$\leftrightarrow$German, WMT16 English$\leftrightarrow$Romanian (EN$\leftrightarrow$RO), and IWSLT14/16 English$\leftrightarrow$German translation benchmark datasets. The decoding iterations and the speedup ratio compared to AT models are reported in Table~\ref{tab:results}.
\begin{table*}[th!]
\centering
\scriptsize
\caption{Performances on popular datasets, i.e., WMT'16 EN$\leftrightarrow$RO, WMT'14 EN$\leftrightarrow$DE, and IWSLT'14/16 EN$\leftrightarrow$DE. ``*'' indicates training with sequence-level knowledge distillation from a big Transformer; ``$\dagger$'' denotes training without sequence-level knowledge distillation; ``$\ddagger$'' refers to results on IWSLT'16.}
    \begin{tabular}{|c|c|c|cc|cc|cc|}
    \hline
   	\multirow{2}{*}{\textbf{Model}} & \multirow{2}{*}{\textbf{Iteration}} & \multirow{2}{*}{\textbf{Speedup}} & \multicolumn{2}{c|}{\textbf{WMT'14}} & \multicolumn{2}{c|}{\textbf{WMT'16}} & \multicolumn{2}{c|}{\textbf{IWSLT'14/16}} \\\cline{4-9}
    & & & \textbf{EN$\to$DE} & \textbf{DE$\to$EN} & \textbf{EN$\to$RO} & \textbf{RO$\to$EN} & \textbf{EN$\to$DE} & \textbf{DE$\to$EN} \\
    \cline{1-9}
    NAT\cite{gu2018non}& 1 & 15.6x & 17.69 & 21.47  & 27.29 & 29.06  & 26.52$\ddagger$ & -\\ \cline{1-9}
    NAT-IR~\cite{lee2018deterministic} & 10 & 1.5x & 21.61  & 25.48 &  29.32 & 30.19 & 27.11$\ddagger$ &32.31$\ddagger$ \\ \cline{1-9}
    RDP~\cite{ding2020understanding} & 2.5 & 3.5x  & 27.8* & - & - & 33.8 & - & - \\ \cline{1-9}
    LRF~\cite{ding2021rejuvenating} & 2.5 & 3.5x & 28.2* & - & - & 33.8 & - & -\\ \cline{1-9}
    SDMRT~\cite{guo2021self} & 10 & - & 27.72* & 31.65* & 33.72 & 33.94 & 27.49 & - \\ \cline{1-9}
    MD~\cite{zhou2020improving} & 1 & - & 25.73 & 30.18 & 31.96 & 33.57 & - & - \\ \cline{1-9}
    DDRS~\cite{shao2022one} & 1 & 14.7x & 27.60 & 31.48 & 34.60 & 
    34.65 & - & 33.12 \\ \cline{1-9}
    LaNMT-C~\cite{zhu2022non} & 2 & 11.0x & 26.02 & 31.23 & 32.50 & - & - & - \\ \cline{1-9}
    CCMLM~\cite{anonymouscontrastive2022} & 10 & - & 27.93* & 31.57* &33.88 & 34.18 & - & - \\ \cline{1-9}
    perLDPE~\cite{oka2021using} & Adaptive & - & 26.3 & 29.5 & - & - & - & - \\ \cline{1-9}
    GLAT~\cite{qian2020glancing}  & 1 & 15.3x & 25.21 & 29.84 &31.19 &32.04 &- &29.61$\dagger$\\ \cline{1-9}  
    PMG~\cite{ding2021progressive} & 2.5 & 3.5x  & 27.8* & - & -  & 33.8* & - & - \\ \cline{1-9}
    \textit{latent}-GLAT~\cite{bao2022glat} & 1  & 11.3x & 26.64 & 29.93 & - & -& -& 32.47    \\ \cline{1-9} 
    Insertion Transformer~\cite{stern2019insertion}& $\approx log_2(N)$ & - & 27.41 & - & - &- & - &-\\ \cline{1-9}
    LevT~\cite{gu2019levenshtein} & Adaptive & 4.0x & 27.27 & - & - & 33.26 &-&-\\ \cline{1-9}  
    CMLM~\cite{ghazvininejad2019mask} & 10 & 1.7x & 27.03* & 30.53* & 33.08 & 33.31 & - & - \\ \cline{1-9}
    SMART~\cite{ghazvininejad2020semi} & 10 & 1.7x & 27.65* & 31.27* & - & - & - & - \\ \cline{1-9}
    DisCo~\cite{kasai2020non} & Adaptive & 3.5x & 27.34* & 31.31* & 33.22 & 33.25 & - & -\\ \cline{1-9} 
    JM-NAT~\cite{guo2020jointly}  & 10 & 5.7x & 27.69* & 32.24 * & 33.52 & 33.72 &- &32.59 \\ \cline{1-9}    
    AR Deep-Shallow~\cite{kasai2020deep} & $N$ & 2.5x & 28.3* & 31.8* & 33.8 & 34.8 & - & -\\ \cline{1-9}
    MvSR-NAT~\cite{xie2022mvsr} & 10 & 3.8x & 27.39* &31.18* &33.38 &33.56  & - &32.55 \\ \cline{1-9}
    R\scriptsize{EWRITE}\footnotesize{NAT}~\cite{geng2021learning} & 2.3 & 3.9x & 27.83* & 31.52* & 33.63 &34.09 &- & -\\ \cline{1-9} 
    CMLMC~\cite{huang2022improving} & 10 & - & 28.37* & 31.41* & 34.57 &34.13 & 28.51 & 34.78\\\cline{1-9} 
    FlowSeq~\cite{ma2019flowseq} & 1  & 1.1x & 23.72 & 28.39 & 29.73 & 30.72 &27.55 &-\\ \cline{1-9}   
    NART-DCRF~\cite{sun2019fast} & 1 & 10.4x & 23.44 & 27.22 & - & - & - & 27.44\\ \cline{1-9}  
    PNAT~\cite{bao2019non} & 1 & 7.3x & 23.05 & 27.18 & - & - & - & 31.23$\ddagger$ \\ \cline{1-9}  
    SynST~\cite{akoury2019syntactically} & $N/6$ & 4.6x & 20.74 & 25.50 & -& - & 23.82 & -\\ \cline{1-9} 
    LaNMT~\cite{shu2020latent} & 1 & 6.8x & 25.10 & - & - & - & - & - \\ \cline{1-9} 
    Imputer~\cite{saharia2020non} & 8 & 3.9x & 28.2*  & 31.8* & 34.4 & 34.1 & - & -\\ \cline{1-9}   
    LAT~\cite{kong2020incorporating}& 4 & 6.7x & 27.35 & 32.04 &32.87 & 33.26 &- & 34.08\\ \cline{1-9}  
    SUNDAE~\cite{savinovstep} & 16 & - & 28.46* & 32.30 * & - & - & - & - \\ \cline{1-9} 
    INSNET~\cite{lu2021efficient} & 16.1 & 3.78x & 28.05 & - & - & 33.91 & - & - \\ \cline{1-9}
    AligNART~\cite{song2021alignart} & 1 &13.2x & 26.4 & 30.4 & 32.5 & 33.1  & - &- \\ \cline{1-9}  
    ReorderNAT~\cite{ran2021guiding}& 1 & 6.0× & 22.79 & 27.28 & 29.30 & 29.50 & 25.29$\ddagger$ & -\\ \cline{1-9}    
    CNAT~\cite{bao2021non} & 1 & 10.4x & 25.56* & 29.36* & - & - & - & 31.15\\ \cline{1-9}   
    SNAT~\cite{liu2021enriching} & 1 & 22.6x &24.64*  & 28.42* &  32.87 & 32.21 & -& -\\ \cline{1-9}   
    Fully NAT~\cite{gu2020fully}& 1 & 16.5x & 27.49 & 31.39 & 33.79 & 34.16 & -& -\\ \cline{1-9}  
    ENAT~\cite{guo2019non} & 1 & 25.3x & 20.65 & 23.02 & 30.08 & - & - & 24.13\\ \cline{1-9}    
    NAT-REG~\cite{wang2019non} & 1 & 27.6x & 20.65 & 24.77 & - & - & 23.14$\ddagger$ & 23.89 \\ \cline{1-9}    
    LAVA NAT~\cite{li2020lava} & 1 & 20.2x & 27.94 & 31.33 & - & 32.85 & - & 33.59$\dagger$ \\ \cline{1-9}    
    CCAN~\cite{ding2020context} & 10 & - & 27.5* & - & - & 33.7 & - & -\\ \cline{1-9}  
    DSLP~\cite{huang2022non}& 1 & 14.8x & 27.02 & 31.61 & 34.17 & 34.60 & - & - \\ \cline{1-9} 
    DAD~\cite{zhan2022non} & 1 & 14.7× & 27.51  & 31.96 & 34.68 & 34.98 &- & - \\ \cline{1-9} 
    DA-Transformer~\cite{huang2022directed} & 1 & 13.9x & 27.49 & 31.37 & - & - & - & - \\ \cline{1-9}
    DA-Transformer Viterbi~\cite{shao2022viterbi} & 1 & 13.2x & 26.89 & 31.10 & - & - & -& - \\ \cline{1-9}
    FA-DAT~\cite{ma2023fuzzy} & 1 & 14.0x & 27.53 & 31.37 & - & - & -& - \\ \cline{1-9} 
    CTC~\cite{saharia2020non}& 1 & 18.6× & 25.7  & 28.10& 32.20 & 31.60 &- & -\\ \cline{1-9}
    Reinforce-NAT~\cite{shao2019retrieving} & 1 & 3.6x & 22.27 & 27.25  & 30.57 & 30.83 & 27.78$\ddagger$& -\\ \cline{1-9} 
    BoN~\cite{shao2020minimizing} & 1 & 9.6x & 20.90 & 24.61 & 28.31 & 29.29 & 25.72$\ddagger$ & -\\ \cline{1-9}  
    AXE~\cite{ghazvininejad2020aligned}& 1 & 15.3x & 23.53* & 27.90* & 30.75 & 31.54 & - & -\\ \cline{1-9}
    O\scriptsize{A}\footnotesize{XE}~\cite{du2021order} & 1 & 15.3x & 26.10* & 30.20* & 32.40 & 33.30 & - & - \\ \cline{1-9} 
    \textit{ngram}-O\scriptsize{A}\footnotesize{XE}~\cite{du2022ngram} & 1 & 15.2x & 26.50* & 30.50* & - & -& - & - \\ \cline{1-9}  
    CoCO~\cite{zhang2022study} & 1 & 14.2x & 27.41 & 31.37 & 34.32 & - & - & - \\ \cline{1-9} 
    MgMO~\cite{li2022multi} & 1 & - & 26.4 & 30.3 & 32.9 & 33.6 & -& - \\ \cline{1-9}
    NMLA~\cite{shaonon} & 1 & 14.7x & 27.57 & 31.28 & 33.86 & 33.94 & - & - \\ \cline{1-9}
    SAT~\cite{wang2018semi}& $N/2$ & 1.5x  & 26.90 & - & - & - & - & -  \\ \cline{1-9} 
    RecoverSAT~\cite{ran2020learning}& $N/2$ & 2.1x & 27.11  & 31.67 & 32.92 & 33.19 & 30.78$\ddagger$ & - \\ \cline{1-9}
    GAD++~\cite{xia2022lossless} & 4.0 & 3.2x & 28.89* & - & - & - & - & - \\ \cline{1-9}
    Unified~\cite{tian2020train} & 10 & - &26.24 & - & - & - & - & 30.73 \\ \cline{1-9} 
    Diformer~\cite{wang2022diformer} & 10 & - & 27.99 & 31.68 & 34.37 & 33.34 & - & - \\ \cline{1-9} 
    HRT~\cite{wang2022hybrid} & $N/2 + 1$ & - & 28.49* & 32.28* & 34.24 & 34.35 & - & - \\ \cline{1-9}
    imitate-NAT~\cite{wei2019imitation}& 1 & 18.4x & 22.44* & 25.67* & 28.61 &28.90 & 28.41$\ddagger$ &-  \\ \cline{1-9} 
    NAT-HINT~\cite{li2019hint}& 1 & 30.2x & 21.11  & 25.24 & - & - & - & 25.55 \\ \cline{1-9}  
    ENGINE~\cite{tu2020engine} & 10 & - & - & - & - & 34.04 & - & 33.17 \\ \cline{1-9}
    EM+ODD~\cite{sun2020approach} & 1 & 16.4x & 24.54  & 27.93 & - & - & - & 30.69 \\ \cline{1-9}  
    FCL-NAT~\cite{guo2020fine}& 1 & 28.9x & 21.70 & 25.32 & - & - & - & 26.62 \\ \cline{1-9} 
    MULTI-TASK NAT~\cite{hao2021multi} & 10 & - & 27.98* & 31.27* & 33.80 & 33.60 & - & - \\ \cline{1-9} 
    TCT-NAT~\cite{liu2020task}& 1 & 27.6x & 21.94 & 25.62 & - & - & 26.01$\ddagger$ & 28.16 \\ \cline{1-9} 
    weak MTL~\cite{wang2022helping} & 1 & - & 27.25 & 30.70 & 33.88 & 34.73 & 35.15 & - \\ \cline{1-9}
    AB-Net~\cite{guo2020incorporating}& - & 2.4x & 28.69* & 33.57* & - & 35.63 & - & 36.49\\ \cline{1-9} 
    NAG-BERT~\cite{su2021non} & 1 & 11.3x & - & - & - & - & - & 30.45 \\ \cline{1-9} 
    CeMAT~\cite{li2022universal}&10 & - & 27.2 & 29.9 & 33.3$\dagger$ & 33.0$\dagger$ & 26.7$\dagger$ & 33.7$\dagger$ \\ \hline
    XML-D~\cite{wang2022xlm} & 8 & 2.8x & 29.80 & 32.88 & 35.34 & 35.50 & - & - \\
    \hline
    \end{tabular}
\label{tab:results}
\end{table*}
Figure \ref{fig:bleuvsspeed} plots the BLEU-Speedup curve to demonstrate the correlations between performance and inference speed achieved by representative NAT methods better, and Figure~\ref{fig:frmework} further presents the evolution of BLEU scores on WMT14 English$\to$German translation by the time of Fully NAT and Iterative NAT.
Methods in the lower left part of Figure~\ref{fig:bleuvsspeed}, e.g., DisCo~\cite{kasai2020non}, NAT~\cite{gu2018non} can achieve much faster inference speed but at the cost of significant performance decrease, while methods in the upper right part can make a better trade-off between speed-up and performance. 
A few powerful NAT methods can even achieve comparable and slightly better performance than the strong AT model with a speed advantage. 
In Figure~\ref{fig:frmework}, iteration-based NAT models generally achieve higher BLEU scores than fully NAT methods at the cost of multiple inference time, but their performance gap is rapidly shrinking, e.g., the recent combination of CTC length prediction, latent variable, and extra upsampling module can achieve competitive performance with strong iteration-based NAT methods.
It can be expected that fully NAT methods can achieve better performance while maintaining their speed advantage with emerging effective strategies and a suitable combination.

\section*{Resources}
We collect valuable resources for NAT models with their open-source information, including the paper URL, code address (Github), and deep learning tools. Table~\ref{tab:url1}, Table~\ref{tab:url2} and Table~\ref{tab:url3} are the summarized information for the resources of NAT task and other extensive tasks.

\begin{table*}[htb]
    \centering
    \scriptsize
    \caption{A collection of NAT published papers and codes.}
    \label{tab:url1}
    \resizebox{\linewidth}{!}{
        \begin{tabular}{ | c | l | l | c |}
        \hline
        Method & Paper URL & Code URL & Framework\\ \hline
        \textbf{Machine Translation}& & & \\\cline{1-4}
        NAT\cite{gu2018non}&  \url{https://openreview.net/pdf?id=B1l8BtlCb} & \url{https://github.com/salesforce/nonauto-nmt} & Pytorch  \\ \cline{1-4}
        NAT-IR~\cite{lee2018deterministic} &  \url{https://aclanthology.org/D18-1149.pdf} & \url{https://github.com/nyu-dl/dl4mt-nonauto} & Pytorch  \\ \cline{1-4}
        RDP~\cite{ding2020understanding} &\url{https://openreview.net/pdf?id=ZTFeSBIX9C} & - & - \\ \cline{1-4}
        LRF~\cite{ding2021rejuvenating} &\url{https://aclanthology.org/2021.acl-long.266.pdf} & \url{https://github.com/longyuewangdcu/RLFW-NAT}  & To be released \\ \cline{1-4}
        SDMRT~\cite{guo2021self} &\url{https://arxiv.org/pdf/2112.11640v1.pdf}  & - & - \\ \cline{1-4}
        MD~\cite{zhou2020improving} & \url{https://aclanthology.org/2020.acl-main.171.pdf}  & - & - \\ \cline{1-4}
        DDRS~\cite{shao2022one} & \url{https://aclanthology.org/2022.naacl-main.277.pdf} & \url{https://github.com/ictnlp/DDRS-NAT} & Pytorch/Fairseq \\ \cline{1-4}
        LaNMT-C~\cite{zhu2022non} & \url{https://aclanthology.org/2022.naacl-main.45.pdf} & \url{https://github.com/zomux/lanmt} & Pytorch \\ \cline{1-4}
        CCMLM~\cite{anonymouscontrastive2022} & \url{https://aclanthology.org/2022.findings-emnlp.463.pdf} & - & - \\ \cline{1-4}
        perLDPE~\cite{oka2021using} & \url{https://arxiv.org/pdf/2107.13689.pdf} & - & - \\ \cline{1-4}        
        GLAT~\cite{qian2020glancing}  &\url{https://aclanthology.org/2021.acl-long.155.pdf} & \url{https://github.com/FLC777/GLAT} & Pytorch/Fairseq \\ \cline{1-4}
        PMG~\cite{ding2021progressive} &\url{https://aclanthology.org/2021.findings-acl.247.pdf} & - & - \\ \cline{1-4}
        \textit{latent}-GLAT~\cite{bao2022glat} & \url{https://aclanthology.org/2022.acl-long.575/} & \url{https://github.com/baoy-nlp/Latent-GLAT} & Pytorch \\ \cline{1-4} 
        Insertion Transformer~\cite{stern2019insertion}&\url{http://proceedings.mlr.press/v97/stern19a/stern19a.pdf} & \url{https://github.com/pytorch/fairseq}& Pytorch/Fairseq\\ \cline{1-4}
        LevT~\cite{gu2019levenshtein} &\url{https://dl.acm.org/doi/pdf/10.5555/3454287.3455290} & \url{https://github.com/pytorch/fairseq} & Pytorch/Fairseq\\ \cline{1-4}  
        CMLM~\cite{ghazvininejad2019mask} &\url{https://aclanthology.org/D19-1633.pdf} & \url{https://github.com/facebookresearch/Mask-Predict} & Pytorch/Fairseq\\ \cline{1-4}  
        SMART~\cite{ghazvininejad2020semi} &\url{https://arxiv.org/pdf/2001.08785.pdf} & - & - \\ \cline{1-4}
        DisCo~\cite{kasai2020non} &\url{http://proceedings.mlr.press/v119/kasai20a/kasai20a.pdf} & \url{https://github.com/facebookresearch/DisCo}  & Pytorch/Fairseq\\ \cline{1-4} 
        JM-NAT~\cite{guo2020jointly}  &\url{https://aclanthology.org/2020.acl-main.36.pdf} &  \url{https://github.com/lemmonation/jm-nat}  & Pytorch/Fairseq\\ \cline{1-4}
        AR Deep-Shallow~\cite{kasai2020deep} &\url{https://openreview.net/pdf?id=KpfasTaLUpq} & \url{https://github.com/jungokasai/deep-shallow} & Pytorch/Fairseq \\ \cline{1-4}
        MvSR-NAT~\cite{xie2022mvsr} &\url{https://ieeexplore.ieee.org/abstract/document/9944912} & - & -\\ \cline{1-4}
        R\scriptsize{EWRITE}\footnotesize{NAT}~\cite{geng2021learning} &\url{https://aclanthology.org/2021.emnlp-main.265.pdf} & \url{https://github.com/xwgeng/RewriteNAT} & Pytorch/Fairseq \\ \cline{1-4} 
        CMLMC~\cite{huang2022improving} &\url{https://openreview.net/pdf?id=I2Hw58KHp8O}&-&-\\ \cline{1-4} 
        FlowSeq~\cite{ma2019flowseq} &\url{https://aclanthology.org/D19-1437.pdf} & \url{https://github.com/XuezheMax/flowseq} & Pytorch \\ \cline{1-4}
        NART-DCRF~\cite{sun2019fast} &\url{https://dl.acm.org/doi/pdf/10.5555/3454287.3454558} & -& -\\ \cline{1-4}  PNAT~\cite{bao2019non} &\url{https://arxiv.org/pdf/1911.10677.pdf} &-&- \\ \cline{1-4}
        SynST~\cite{akoury2019syntactically} &\url{https://aclanthology.org/P19-1122.pdf} & \url{https://github.com/dojoteef/synst}& Pytorch\\ \cline{1-4} 
        LaNMT~\cite{shu2020latent} &\url{https://ojs.aaai.org/index.php/AAAI/article/view/6413} &  \url{https://github.com/zomux/lanmt} & Pytorch  \\ \cline{1-4} 
        Imputer~\cite{saharia2020non} &\url{https://aclanthology.org/2020.emnlp-main.83.pdf} &\url{https://github.com/rosinality/imputer-pytorch} & Pytorch  \\ \cline{1-4} 
        LAT~\cite{zhuincorporating}&\url{https://aclanthology.org/2020.emnlp-main.79.pdf} & \url{https://github.com/shawnkx/NAT-with-Local-AT}& Pytorch\\ \cline{1-4} 
        AligNART~\cite{song2021alignart} &\url{https://aclanthology.org/2021.emnlp-main.1.pdf} &-&- \\ \cline{1-4} 
        ReorderNAT~\cite{ran2021guiding}&\url{https://ojs.aaai.org/index.php/AAAI/article/view/17618} & \url{https://github.com/ranqiu92/ReorderNAT} & Pytorch/OpenNMT \\ \cline{1-4} 
        CNAT~\cite{bao2021non} &\url{https://aclanthology.org/2021.naacl-main.458.pdf} & \url{https://github.com/baoy-nlp/CNAT}  & Pytorch \\ \cline{1-4} 
        SNAT~\cite{liu2021enriching} &\url{https://aclanthology.org/2021.eacl-main.105.pdf} &-&- \\ \cline{1-4} 
        Fully NAT~\cite{gu2020fully}&\url{https://aclanthology.org/2021.findings-acl.11.pdf} & \url{https://github.com/pytorch/fairseq}& Pytorch/Fairseq \\ \cline{1-4}  
        ENAT~\cite{guo2019non} &\url{https://ojs.aaai.org/index.php/AAAI/article/view/4257} &-&- \\ \cline{1-4} 
        NAT-REG~\cite{wang2019non} &\url{https://ojs.aaai.org/index.php/AAAI/article/view/4476} &-&- \\ \cline{1-4} 
        LAVA NAT~\cite{li2020lava} &\url{https://arxiv.org/pdf/2002.03084v1.pdf} &-&-\\ \cline{1-4} 
        CCAN~\cite{ding2020context} &\url{https://aclanthology.org/2020.coling-main.389.pdf} &-&- \\ \cline{1-4} 
        DSLP~\cite{huang2022non}&\url{https://ojs.aaai.org/index.php/AAAI/article/view/21323}&\url{https://github.com/chenyangh/DSLP} & Pytorch/Fairseq \\ \cline{1-4}
        DAD~\cite{zhan2022non} &\url{https://arxiv.org/pdf/2203.16266.pdf} & \url{ https://github.com/zja-nlp/NAT_with_DAD} & Pytorch/Fairseq\\ \cline{1-4} 
        CTC~\cite{graves2006connectionist}&\url{https://www.cs.toronto.edu/~graves/icml_2006.pdf} & \url{https://github.com/parlance/ctcdecode} & C++ \\ \cline{1-4}
        Reinforce-NAT~\cite{shao2019retrieving} &\url{https://aclanthology.org/P19-1288.pdf} & \url{https://github.com/ictnlp/RSI-NAT} & Pytorch \\ \cline{1-4} 
        BoN~\cite{shao2020minimizing} &\url{https://ojs.aaai.org/index.php/AAAI/article/view/5351} & \url{https://github.com/ictnlp/BoN-NAT}& Fairseq\\ \cline{1-4}  
        AXE~\cite{ghazvininejad2020aligned}&\url{http://proceedings.mlr.press/v119/ghazvininejad20a/ghazvininejad20a.pdf} & \url{https://github.com/m3yrin/aligned-cross-entropy} & Pytorch\\ \cline{1-4}
        EISL~\cite{liu2022don}&\url{https://aclanthology.org/2022.naacl-main.150.pdf} & \url{https://github.com/guangyliu/EISL} & Pytorch/Fairseq \\ \cline{1-4} 
        O\scriptsize{A}\footnotesize{XE}~\cite{du2021order} &\url{http://proceedings.mlr.press/v139/du21c/du21c.pdf} & \url{https://github.com/tencent-ailab/ICML21_OAXE}& Pytorch/Fairseq\\ \cline{1-4} 
        SAT~\cite{wang2018semi}&\url{https://aclanthology.org/D18-1044.pdf}  &-&-\\ \cline{1-4} 
        
        SUNDAE~\cite{savinovstep}& \url{https://openreview.net/pdf?id=T0GpzBQ1Fg6} & \url{https://github.com/vvvm23/sundae} & Pytorch \\ \cline{1-4} 
        INSNET~\cite{lu2021efficient}& \url{https://openreview.net/pdf?id=vsShetzoRG9} & - & -  \\ \cline{1-4} 
        DA-Transformer~\cite{huang2022directed}& \url{https://proceedings.mlr.press/v162/huang22m/huang22m.pdf} & \url{https://github.com/thu-coai/DA-Transformer} & Pytorch/Fairseq \\ \cline{1-4} 
        DA-Transformer Viterbi~\cite{shao2022viterbi}& \url{https://aclanthology.org/2022.findings-emnlp.322.pdf} &- & -\\ \cline{1-4} 
        FA-DAT~\cite{ma2023fuzzy} & \url{https://openreview.net/pdf?id=LSz-gQyd0zE} & - & - \\ \cline{1-4}
        \textit{ngram}-O\scriptsize{A}\footnotesize{XE}~\cite{du2022ngram}& \url{https://aclanthology.org/2022.coling-1.446.pdf} & - & -\\ \cline{1-4} 
        CoCO~\cite{zhang2022study}& \url{https://aclanthology.org/2022.naacl-main.126.pdf} & - & -\\ \cline{1-4} 
        MgMO~\cite{li2022multi}& \url{https://aclanthology.org/2022.emnlp-main.339.pdf} & - & - \\ \cline{1-4} 
        NMLA~\cite{shaonon}& \url{https://openreview.net/pdf?id=Qvh0SAPrYzH} & \url{https://github.com/ictnlp/NMLA-NAT} & Pytorch/Fairseq \\ \cline{1-4} 
        GAD++~\cite{xia2022lossless}& \url{https://arxiv.org/pdf/2203.16487v2.pdf} & \url{https://github.com/hemingkx/Generalized-Aggressive-Decoding} & Pytorch/Fairseq \\ \cline{1-4} 
        HRT~\cite{wang2022hybrid}& \url{https://openreview.net/pdf?id=2NQ8wlmU9a_} & - & -\\ \cline{1-4} 
        weak MTL~\cite{wang2022helping}& \url{https://aclanthology.org/2022.emnlp-main.371.pdf} & \url{https://github.com/wxy-nlp/MultiTaskNAT} & - \\ \cline{1-4}      RecoverSAT~\cite{ran2020learning}&\url{https://aclanthology.org/2020.acl-main.277.pdf} &  \url{https://github.com/ranqiu92/RecoverSAT} & Pytorch/OpenNMT \\ \cline{1-4}
        Unified~\cite{tian2020train} & \url{https://aclanthology.org/2020.coling-main.25.pdf} & - & -\\ \cline{1-4} 
        Diformer~\cite{wang2022diformer} &\url{https://aclanthology.org/2022.eamt-1.11.pdf} & - & -\\ \cline{1-4} 
        imitate-NAT~\cite{wei2019imitation}&\url{https://aclanthology.org/P19-1125.pdf}  &-&-\\ \cline{1-4} 
        NAT-HINT~\cite{li2019hint}&\url{https://aclanthology.org/D19-1573.pdf} & \url{https://github.com/zhuohan123/hint-nart}  & Pytorch\\ \cline{1-4} 
        ENGINE~\cite{tu2020engine} &\url{https://aclanthology.org/2020.acl-main.251.pdf} & \url{https://github.com/lifu-tu/ENGINE} & Pytorch/Fairseq\\ \cline{1-4}
        EM+ODD~\cite{sun2020approach} &\url{http://proceedings.mlr.press/v119/sun20c/sun20c.pdf} & \url{https://github.com/Edward-Sun/NAT-EM} & Pytorch\\ \cline{1-4}  
        FCL-NAT~\cite{guo2020fine}&\url{https://ojs.aaai.org/index.php/AAAI/article/view/6289} & \url{https://github.com/lemmonation/fcl-nat}  & Tensorflow/Tensortotensor \\ \cline{1-4} 
        MULTI-TASK NAT~\cite{hao2021multi} &\url{https://aclanthology.org/2021.naacl-main.313.pdf} & \url{https://github.com/yongchanghao/multi-task-nat} & Pytorch/Fairseq \\ \cline{1-4} 
        TCT-NAT~\cite{liu2020task}&\url{https://www.ijcai.org/Proceedings/2020/0534.pdf} &-&- \\ \cline{1-4}
        AB-Net~\cite{guo2020incorporating}&\url{https://dl.acm.org/doi/pdf/10.5555/3495724.3496634} & \url{https://github.com/lemmonation/abnet}& Pytorch/Fairseq\\ \cline{1-4} 
        NAG-BERT~\cite{su2021non} &\url{https://aclanthology.org/2021.eacl-main.18.pdf} & \url{https://github.com/yxuansu/NAG-BERT}& Pytorch/Fairseq \\ \cline{1-4} 
        CeMAT~\cite{li2022universal}&\url{https://aclanthology.org/2022.acl-long.442.pdf} & \url{https://github.com/huawei-noah}& Pytorch/Fairseq \\ \cline{1-4}
        XML-D~\cite{wang2022xlm} &\url{https://aclanthology.org/2022.emnlp-main.466.pdf} & - & - \\ \cline{1-4}
        
    \end{tabular}}
\end{table*}

\begin{table*}[htb]
    \centering
    \scriptsize
    \caption{A collection of NAR related published papers and codes on general-purpose and task specific text generation tasks.  }\label{tab:url2}
    \resizebox{\linewidth}{!}{
        \begin{tabular}{ | c | l | l | c |}
        \hline
        Method & Paper URL & Code URL & Framework\\ \hline
        \textbf{General-Purpose}& & & \\
        \textbf{Text Generation}& & & \\\cline{1-4}
        POSPD~\cite{yang2021pos} & \url{https://aclanthology.org/2021.acl-long.467.pdf} & \url{https://github.com/yangkexin/pospd} & Pytorch/Fairseq\\ \cline{1-4}   MIST~\cite{jiang2021improving}&\url{https://arxiv.org/pdf/2110.11115v1.pdf} & \url{https://github.com/kongds/mist} & Pytorch/Fairseq\\ \cline{1-4}    BANG~\cite{qi2021bang}&\url{http://proceedings.mlr.press/v139/qi21a/qi21a.pdf}  &\url{https://github.com/microsoft/BANG}& - \\ \cline{1-4}
        EDITCL~\cite{agrawal2022imitation} & \url{https://aclanthology.org/2022.acl-long.520.pdf} & - & - \\ \cline{1-4}
        NAG-BERT~\cite{su2021non} & \url{https://aclanthology.org/2021.eacl-main.18.pdf} & \url{https://github.com/yxuansu/NAG-BERT} & Pytorch \\ \cline{1-4}
        EDIT5~\cite{mallinson2022edit5} &  \url{https://arxiv.org/pdf/2205.12209.pdf} & - & - \\ \cline{1-4} 
        ELMER~\cite{li2022elmer} & \url{https://aclanthology.org/2022.emnlp-main.68.pdf} & \url{https://github.com/RUCAIBox/ELMER} & Pytorch/Fairseq \\ \cline{1-4}
        
        \textbf{Summarization} & & & \\ \cline{1-4}
        NAUS~\cite{liu2022learning} & \url{https://aclanthology.org/2022.acl-long.545.pdf} & - & - \\ \cline{1-4} 
        NACC~\cite{liucharacter} & \url{https://openreview.net/pdf?id=KXybrIUJnya} & \url{https://github.com/MANGA-UOFA/NACC} & Pytorch/Fairseq \\ \cline{1-4}
        \textbf{Dialogue} & & & \\ \cline{1-4} 
        CG-nAR~\cite{zou2021thinking}&\url{https://aclanthology.org/2021.emnlp-main.169.pdf}  &\url{https://github.com/rowitzou/cg-nar} & Pytorch/Transformers\\ \cline{1-4}
        NonAR+MMI~\cite{han2020non}&\url{https://arxiv.org/pdf/2002.04250v2.pdf} & -&- \\ \cline{1-4}
        GL-GIN~\cite{qin2021gl}&\url{https://aclanthology.org/2021.acl-long.15.pdf} &\url{https://github.com/yizhen20133868/GL-GIN} & Pytorch \\ \cline{1-4}
        SlotRefine~\cite{wu2020slotrefine} &\url{https://aclanthology.org/2020.emnlp-main.152.pdf}  &\url{https://github.com/moore3930/SlotRefine} & Tensorflow  \\ \cline{1-4}
        NADST~\cite{le2020non}&\url{https://openreview.net/pdf?id=H1e_cC4twS} & \url{https://github.com/henryhungle/NADST} & PyTorch \\ \cline{1-4}
        LR-Transformer~\cite{cheng2021effective} & \url{https://dl.acm.org/doi/abs/10.1145/3459637.3482229} &- & - \\  \cline{1-4}
        \textbf{Grammatical Error} & & & \\ 
        \textbf{Correction} & & & \\ \cline{1-4}
        TtT~\cite{li2021tail}&\url{https://aclanthology.org/2021.acl-long.385.pdf}  &\url{https://github.com/lipiji/TtT}& Pytorch\\ \cline{1-4}
        BERT-GEC~\cite{straka2021character} &\url{https://aclanthology.org/2021.wnut-1.46.pdf}  &\url{https://github.com/ufal} & Pytorch \\ \cline{1-4}
        MaskCorrect~\cite{shen2022mask} & \url{https://arxiv.org/pdf/2211.13252.pdf}
        & - & - \\ \cline{1-4}
        \textbf{Text Style Transfer} & & & \\ \cline{1-4}
        NAST~\cite{huang2021nast}. & \url{https://aclanthology.org/2021.findings-acl.138.pdf} & \url{https://github.com/thu-coai/NAST} &-\\  \cline{1-4}
        KD+CL+ID~\cite{ma2021exploring} & \url{https://aclanthology.org/2021.emnlp-main.730.pdf} &  \url{https://github.com/sunlight-ym/nar_style_transfer} &-\\  \cline{1-4}
        \textbf{Controllable Text} & & & \\ 
        \textbf{Generation} & & & \\ \cline{1-4}
        PMI~\cite{agrawal2021non} & \url{https://aclanthology.org/2021.findings-acl.330.pdf} & - & - \\ \cline{1-4} 
        MUCOLA~\cite{kumar2022gradient} & \url{https://aclanthology.org/2022.emnlp-main.144.pdf} & \url{https://github.com/Sachin19/mucoco/tree/sampling} & Pytorch \\ \cline{1-4}
        Diffusion-LM~\cite{lidiffusion} & \url{https://openreview.net/pdf?id=3s9IrEsjLyk} & \url{https://github.com/XiangLi1999/Diffusion-LM} & Pytorch/Transformers \\ \cline{1-4}
        AutoTemplate~\cite{iso2022autotemplate} & \url{https://arxiv.org/pdf/2211.08387.pdf} & - & - \\ \cline{1-4}
        \textbf{Image} & & & \\ 
        \textbf{Captioning} & & & \\ \cline{1-4}
        TIger~\cite{wang2022explicit} & \url{https://link.springer.com/chapter/10.1007/978-3-031-20059-5_7} & \url{https://github.com/baaaad/ECE} & Pytorch \\ \cline{1-4}
        FutureCap~\cite{fei2022efficient} & \url{https://dl.acm.org/doi/abs/10.1145/3503161.3547840} & \url{https://github.com/feizc/Future-Caption} & Pytorch/Transformers \\ \cline{1-4}
        Transformer-DML~\cite{chenlearning} & \url{https://openreview.net/pdf?id=LMuh9bS4tqF} & \url{https://github.com/bladewaltz1/ModeCap} & Pytorch/Transformers \\ \cline{1-4}
        UAIC~\cite{fei2022uncertainty} & \url{https://arxiv.org/pdf/2211.16769.pdf} & - & - \\ \cline{1-4}
        \textbf{Question} & & & \\ 
        \textbf{Answering} & & & \\  \cline{1-4}
        NAPG~\cite{zhang2022napg} & \url{https://arxiv.org/pdf/2211.03462.pdf} & - & - \\ \cline{1-4}
        KECP~\cite{wang2022kecp} & \url{https://arxiv.org/pdf/2205.03071.pdf} & \url{https://github.com/alibaba/EasyNLP} & Pytorch/EasyNLP \\ \cline{1-4}
        \textbf{Automatic Speech }& & & \\
        \textbf{Recognition}& & & \\\cline{1-4}
        NAR CTC/attention~\cite{deng2022improving} &\url{https://ieeexplore.ieee.org/abstract/document/9746316} & -& -\\  \cline{1-4}
        S-CFE CTC~\cite{komatsu2022non} &\url{https://ieeexplore.ieee.org/abstract/document/9746770} &-& -\\  \cline{1-4}
        CASS-NAT~\cite{fan2021cass}&\url{https://ieeexplore.ieee.org/abstract/document/9413429} &-&- \\  \cline{1-4}
        DLP~\cite{higuchi2021improved}&\url{https://ieeexplore.ieee.org/abstract/document/9414198} &-& -\\  \cline{1-4}
        CTC-enhanced~\cite{song2021non} &\url{https://ieeexplore.ieee.org/abstract/document/9414694} &-&- \\  \cline{1-4}
        Align-Refine~\cite{chi2021align} &\url{https://aclanthology.org/2021.naacl-main.154.pdf} &\url{https://github.com/amazon-research/align-refine} &To be released \\  \cline{1-4}
        Align-Denoise~\cite{chen2021align}&\url{http://dx.doi.org/10.21437/Interspeech.2021-1906}&\url{https://github.com/bobchennan/espnet/tree}&Pytorch/Espnet \\  \cline{1-4}
        LASO~\cite{bai2021fast}&\url{https://ieeexplore.ieee.org/document/9437636} &-& -\\  \cline{1-4}
        NAR-BERT-ASR~\cite{yu2021non}&\url{https://arxiv.org/pdf/2104.04805v1.pdf} &-&- \\  \cline{1-4}
        CondChain~\cite{guo2022multi}&\url{https://arxiv.org/pdf/2106.08595v1.pdf} & \url{https://github.com/pengchengguo/espnet} & Pytorch/Espnet\\  \cline{1-4}
        Streaming NAR~\cite{wang2021streaming}&\url{https://arxiv.org/pdf/2107.09428v1.pdf} & \url{https://github.com/espnet/espnet} & Pytorch/Espnet\\  \cline{1-4}
        Mask-CTC~\cite{higuchi2020mask}&\url{http://www.interspeech2020.org/uploadfile/pdf/Thu-1-3-7.pdf} & \url{https://github.com/espnet/espnet} & Pytorch/Espnet \\  \cline{1-4}
        Intermediate CTC~\cite{lee2021intermediate}& \url{https://ieeexplore.ieee.org/abstract/document/9414594/} & \url{https://github.com/espnet/espnet}& Pytorch/Espnet \\ \cline{1-4}
        Self-Conditioned CTC~\cite{nozaki2021relaxing}&\url{https://arxiv.org/pdf/2104.02724.pdf} & \url{https://github.com/espnet/espnet}& Pytorch/Espnet \\ \cline{1-4}
        GIC~\cite{yang2022improving} & \url{https://arxiv.org/pdf/2205.12462.pdf} & - & - \\ \cline{1-4}
        CAKT~\cite{lu2023context} & \url{https://ieeexplore.ieee.org/abstract/document/10022825} & - & - \\ \cline{1-4}
        Inter-KD~\cite{yoon2023inter} & \url{https://ieeexplore.ieee.org/abstract/document/10022581} & - & - \\ \cline{1-4} 
        BECTRA~\cite{higuchi2022bectra} & \url{https://arxiv.org/pdf/2211.00792.pdf} & - & - \\ \cline{1-4}

    \end{tabular}}
\end{table*}

\begin{table*}[htb]
    \centering
    \scriptsize
    \caption{A collection of other NAR related published papers and codes beyond text generation. IE denotes information extraction task, VG denotes video generation task, VC denotes voice conversion task, and SR denotes speech representation task,
    CC denotes code completion task, PTSF denotes time series forecasting task.  }\label{tab:url3}
    \resizebox{\linewidth}{!}{
        \begin{tabular}{ | c | l | l | c |}
        \hline
        Method & Paper URL & Code URL & Framework\\ \hline
        \textbf{Text to}& & & \\
        \textbf{Speech}& & & \\\cline{1-4}
        BVAE-TTS~\cite{lee2020bidirectional}&\url{https://openreview.net/pdf?id=o3iritJHLfO} & \url{https://github.com/LEEYOONHYUNG/BVAE-TTS} & Pytorch\\ \cline{1-4}
        GAN-TTS~\cite{guo2022multi}& \url{https://arxiv.org/pdf/2203.01080.pdf}& \url{https://github.com/yanggeng1995/GAN-TTS} & Pytorch\\  \cline{1-4}
        VARA-TTS~\cite{liu2021vara}& \url{https://arxiv.org/pdf/2102.06431v1.pdf}& \url{https://github.com/vara-tts/VARA-TTS} & - \\  \cline{1-4}
        Glow-TTS~\cite{kim2020glow} & \url{https://dl.acm.org/doi/pdf/10.5555/3495724.3496400} & \url{https://github.com/jaywalnut310/glow-tts} & Tensorflow/Tensor2tensor \\  \cline{1-4}
        VAENAR-TTS~\cite{lu2021vaenar}& \url{https://arxiv.org/pdf/2107.03298v1.pdf} & \url{https://github.com/thuhcsi/VAENAR-TTS} & Pytorch\\  \cline{1-4}
        ParaNet~\cite{peng2020non} & \url{http://proceedings.mlr.press/v119/peng20a/peng20a.pdf} & \url{https://github.com/ksw0306/WaveVAE}& Pytorch\\  \cline{1-4}
        FastSpeech~\cite{ren2019fastspeech}& \url{https://dl.acm.org/doi/pdf/10.5555/3454287.3454572} & \url{https://github.com/coqui-ai/TTS}& PyTorch/TTS\\  \cline{1-4}
        TalkNet2~\cite{beliaev2021talknet}& \url{https://arxiv.org/pdf/2104.08189v3.pdf} &  \url{https://github.com/rishikksh20/TalkNet2-pytorch} & -\\  \cline{1-4}
        FastSpeech2~\cite{chien2021hierarchical}&
        \url{https://ieeexplore.ieee.org/abstract/document/9383629} & \url{https://github.com/ming024/FastSpeech2}& Pytorch \\  \cline{1-4}
        HiMuV-TTS~\cite{Bae2022hierarchical}& \url{https://arxiv.org/pdf/2204.04004.pdf} & -& - \\ \cline{1-4}
        CLONE~\cite{liu2022controllable} & \url{https://arxiv.org/pdf/2207.06088.pdf} & - & - \\ \cline{1-4}
        CUC-VAE~\cite{li2022cross} & \url{https://aclanthology.org/2022.acl-long.30.pdf} & - & - \\ \cline{1-4} 
        \textbf{Speech}& & & \\
        \textbf{translation}& & & \\\cline{1-4}
        MTL~\cite{chuang2021investigating}& \url{https://aclanthology.org/2021.findings-acl.92.pdf} &  \url{https://github.com/voidism/NAR-ST}& Pytorch/Espnet \\  \cline{1-4}
        Orthros~\cite{inaguma2021orthros}& \url{https://ieeexplore.ieee.org/abstract/document/9415093}& - & -\\  \cline{1-4}
        Orthros-CMLM~\cite{inaguma2021non}& \url{https://arxiv.org/pdf/2109.04411v1.pdf}& - & -\\  \cline{1-4}
        Fast-MD~\cite{inaguma2021fast}& \url{https://ieeexplore.ieee.org/abstract/document/9687894}& - & -\\  \cline{1-4}

        \textbf{Semantic}& & & \\
        \textbf{Parsing}& & & \\\cline{1-4}
        Span Pointer~\cite{shrivastava2021span}&\url{https://aclanthology.org/2021.findings-emnlp.161.pdf} & - & -\\ \cline{1-4}
        LightConv Pointer~\cite{babu2021non} &\url{https://aclanthology.org/2021.naacl-main.236.pdf}  &\url{https://github.com/facebookresearch/pytext}& Pytorch/Pytest\\ \cline{1-4}
        RNGTr~\cite{mohammadshahi2021recursive}&\url{https://aclanthology.org/2021.tacl-1.8.pdf}  &\url{https://github.com/idiap/g2g-transformer} & Pytorch\\ \cline{1-4}

        \textbf{Diffusion} & & & \\
        \textbf{Models} & & & \\ \cline{1-4}
        WaveGrad~\cite{kong2020diffwave} & \url{https://openreview.net/pdf?id=NsMLjcFaO8O} & - & - \\ \cline{1-4}
        DDPM~\cite{ho2020denoising} & \url{https://dl.acm.org/doi/pdf/10.5555/3495724.3496298} & \url{https://arxiv.org/pdf/2006.11239.pdf} & Tensorflow \\ \cline{1-4}
        D3PMs~\cite{austin2021structured} & \url{https://openreview.net/pdf?id=h7-XixPCAL} & - & - \\ \cline{1-4}
        Imagen~\cite{sahariaphotorealistic} & \url{https://openreview.net/pdf?id=08Yk-n5l2Al} & - & - \\ \cline{1-4}
        unCLIP~\cite{ramesh2022hierarchical} & \url{https://arxiv.org/pdf/2204.06125.pdf} & - & - \\ \cline{1-4}
        GLIDE~\cite{nichol2022glide} & \url{https://proceedings.mlr.press/v162/nichol22a/nichol22a.pdf} & \url{https://github.com/openai/glide-text2im} & Pytorch \\ \cline{1-4}
        classifier-free guidance~\cite{ho2021classifier} & \url{https://openreview.net/forum?id=qw8AKxfYbI} & - & - \\ \cline{1-4}
        LDEBM~\cite{yu2022latent} & \url{https://proceedings.mlr.press/v162/yu22h/yu22h.pdf} & - & - \\ \cline{1-4}
        Diffusion-LM~\cite{lidiffusion} & \url{https://openreview.net/pdf?id=3s9IrEsjLyk} & \url{https://github.com/XiangLi1999/Diffusion-LM} & Pytorch/Transformers \\ \cline{1-4}
        DIFFUSER~\cite{reid2022diffuser} & \url{https://arxiv.org/pdf/2210.16886.pdf} & - & - \\ \cline{1-4}
        DIFFUSEQ~\cite{gong2022diffuseq} & \url{https://arxiv.org/pdf/2210.08933.pdf} & \url{https://github.com/Shark-NLP/DiffuSeq} & Pytorch/Transformers \\ \cline{1-4}
        DiffGAR~\cite{yin2022diffgar} & \url{https://arxiv.org/pdf/2210.08573.pdf} & - & - \\ \cline{1-4}
        \textbf{Others}& & & \\\cline{1-4}
        MacroIE~\cite{yu2021maximal}(IE) & \url{https://aclanthology.org/2021.emnlp-main.764.pdf} &- & - \\  \cline{1-4}
        DIGAN~\cite{yu2021generating}(VG)& \url{https://openreview.net/pdf?id=Czsdv-S4-w9}& - & -\\  \cline{1-4}
        FastSpeech2-VC~\cite{hayashi2021non}(VC)& \url{https://ieeexplore.ieee.org/abstract/document/9413973}& - & -\\  \cline{1-4}
        CDHVAE~\cite{akuzawa2021conditional}(VC) & \url{https://ieeexplore.ieee.org/abstract/document/9689369}& - & -\\ \cline{1-4}
        NPC~\cite{liu2020non}(SR)& \url{https://arxiv.org/pdf/2011.00406v1.pdf} &  \url{https://github.com/Alexander-H-Liu/NPC}& Pytorch\\ \cline{1-4}
        SANAR~\cite{liu2022non}(CC) & \url{https://arxiv.org/pdf/2204.09877.pdf} & - & - \\ \cline{1-4}
        MANF~\cite{feng2022multi}(TSF) & \url{https://arxiv.org/pdf/2205.07493.pdf} & - & - \\ \cline{1-4}   
    \end{tabular}}
\end{table*}

\ifCLASSOPTIONcaptionsoff
  \newpage
\fi
\end{document}